\documentclass[twoside]{article}

\usepackage[accepted]{aistats2023}
\usepackage{listings}
\usepackage{natbib}
\usepackage{tabularx} 
\usepackage{amsmath}  
\usepackage{graphicx} 
\usepackage{rotating} 
\usepackage[T1]{fontenc}
\usepackage{aecompl}
\usepackage{docmute} 
\usepackage[margin=1in,letterpaper]{geometry} 
\usepackage[final]{hyperref} 
\hypersetup{
	colorlinks=true,       
	linkcolor=blue,        
	citecolor=blue,        
	filecolor=magenta,     
	urlcolor=blue         
}

\usepackage{bm}
\usepackage{bbding}
\usepackage{amssymb}
\usepackage{color}
\usepackage{algorithm}
\usepackage{algorithmic}
\newlength\myindent
\usepackage{multirow}
\usepackage{makecell}
\usepackage{amsthm}
\usepackage{subfigure}

\usepackage{threeparttable}

\usepackage{caption}

\begin{document}

%
\runningtitle{Efficient Informed Proposals for Discrete Distributions via Newton’s Series Approximation}

%
\runningauthor{Yue Xiang, Dongyao Zhu, Bowen Lei, Dongkuan Xu, Ruqi Zhang}

\twocolumn[

\aistatstitle{Efficient Informed Proposals for Discrete Distributions \\via Newton’s Series Approximation}

\aistatsauthor{Yue Xiang \And Dongyao Zhu \And  Bowen Lei}
\aistatsaddress{Renmin University of China \And Independent Researcher \And Texas A\&M University}

\aistatsauthor{Dongkuan Xu \And Ruqi Zhang }
\aistatsaddress{North Carolina State University \And Purdue University}]

\begin{abstract}
Gradients have been exploited in proposal distributions to accelerate the convergence of Markov chain Monte Carlo algorithms on discrete distributions. However, these methods require a natural differentiable extension of the target discrete distribution, which often does not exist or does not provide effective gradient guidance.
In this paper, we develop a gradient-like proposal for any discrete distribution without this strong requirement. Built upon a locally-balanced proposal, our method efficiently approximates the discrete likelihood ratio via Newton's series expansion to enable a large and efficient exploration in discrete spaces.
We show that our method can also be viewed as a multilinear extension, thus inheriting its desired properties. We prove that our method has a guaranteed convergence rate with or without the Metropolis-Hastings step. Furthermore, our method outperforms a number of popular alternatives in several different experiments, including the facility location problem, extractive text summarization, and image retrieval.
\end{abstract}

\section{INTRODUCTION}
Discrete structures are common in the real world, from discrete data such as text~\citep{wang2019bert, gu2017non} and genomes~\citep{wang2010gibbs}, to discrete models such as low-precision neural networks~\citep{courbariaux2016binarized,peters2018probabilistic} and graphical models of molecules~\citep{gilmer2017neural}. As data and models become complex and large-scale, it is desirable to develop efficient proposals in Markov chain Monte Carlo (MCMC) algorithms that allow us to sample from these complex high-dimensional discrete distributions~\citep{zhang2022langevin}.

Gradients have been widely utilized in proposal distributions to accelerate the convergence of MCMC, such as the Langevin algorithm~\citep{roberts1996exponential,roberts2002langevin} and Hamiltonian Monte Carlo (HMC)~\citep{duane1987hybrid, neal2011mcmc}. These gradient-based methods are mainly designed for continuous distributions and require a natural neighborhood to define gradients.
However, since there is no natural neighborhood in discrete distributions, it becomes challenging to incorporate gradients in the proposal to accelerate sampling.

Previous research has been devoted to making gradient-based proposals for efficient discrete sampling, however, they either require natural differentiable relaxation or sacrifice convergence speed. As shown in the ``Natural continuous
extension available'' column in Table~\ref{table:compare}, Gibbs with gradient proposal~\citep{grathwohl2021oops} and discrete Langevin proposal~(DLP)~\citep{zhang2022langevin} assume the existence of an underlying differentiable distribution in the discrete space and exploit gradient information to speed up sampling and inference. In the top right of Table~\ref{table:compare}, the locally-balanced proposal~\citep{zanella2020informed} does not require this assumption, while it only conducts local moves in small windows, which leads to slow convergence, especially in high-dimensional tasks.
Therefore, the question we need to answer is how to design a method that can sample efficiently in discrete spaces and has no strong requirement of natural continuous expansion.

To answer this question, we present a gradient-like informed proposal to efficiently sample from any discrete distribution without the strong requirement of continuous relaxations.
To find more informative directions during sampling, our method estimates the likelihood ratio of making discrete moves through Newton's series expansion, so we call our proposal \emph{Newton proposal}.
In addition, we design a coordinatewise factorization scheme in our method, so we can update multiple coordinates in a single move, which further improves the sampling efficiency.
We summarize our contributions as follows:
\begin{itemize}
    \item We propose a new informed proposal, Newton proposal, for discrete distributions. It allows multiple coordinates to be updated simultaneously while not requiring that the discrete distribution to be naturally extended to the continuous domain.
    \item We show that our Newton proposal can be obtained from Newton's series approximation to the target discrete distribution or from Taylor expansion to the multilinear extension of the discrete distribution, which justifies Newton proposal's desirable properties.  
    \item We theoretically prove the convergence rate of our Newton scheme without and with the Metropolis-Hastings correction, demonstrating its efficient sampling in discrete distributions.
    \item We experimentally show that Newton proposal outperforms existing discrete proposals and some optimization-based methods when sampling from high-dimensional complex discrete distributions, including facility location, text summarization, and image retrieval.
\end{itemize}

\begin{table}[!t]
\vspace{-0.2cm}
\scriptsize
\centering
\caption{Proposals for discrete distributions.}
\begin{tabular}{l|ll}
\hline
& \makecell[l]{Natural differentiable\\ extension available} 
& \makecell[l]{Natural differentiable\\ extension unavailable}\\ 
\hline
\begin{tabular}[c]{@{}l@{}}Update one\\ coordinate in a step\end{tabular} &
  \begin{tabular}[c]{@{}l@{}}Gibbs with \\gradient proposal\end{tabular} &
  \begin{tabular}[c]{@{}l@{}}Locally-balanced \\ proposal\end{tabular} \\ \hline
\begin{tabular}[c]{@{}l@{}}Update multiple\\ coordinates in a step\end{tabular} &
  \begin{tabular}[c]{@{}l@{}}Discrete Langevin \\ proposal\end{tabular} &
  \begin{tabular}[c]{@{}l@{}}\textbf{Newton proposal (Ours)}\end{tabular} \\ \hline
\end{tabular}
\label{table:compare}
\vspace{-0.6cm}
\end{table}

\section{RELATED WORK}
\textbf{Informed Proposal}
Various informed Metropolis-Hastings (MH) proposal distributions have been designed to avoid slow mixing and slow convergence brought by random walk MH proposals.
Using symmetric proposal distributions, random walk MH schemes are easy to implement, but no information about the target distribution is utilized and the new state is proposed randomly. 
In the contrary, informed proposal distributions elaborate information about the target distribution, such as the gradient of the target to bias the proposal distribution towards high probability, resulting in substantial improvements of MCMC performances. However, most of these informed proposals are based on derivatives and it is nontrivial to extend such methods to discrete spaces. 
As a consequence, most MCMC proposals for discrete spaces often rely on symmetric and uninformed proposal distributions, which can induce slow convergence.

\textbf{Continuous Relaxation-Based Method} 
Gradient-based informed proposals can be applied to discrete distributions via continuous relaxations~\citep{pakman2013auxiliary,nishimura2020discontinuous,han2020stein,zhou2020mixed,jaini2021sampling,zhang2022generative}.
They are usually implemented by transporting the problem into a continuous domain, performing updates under gradient-based proposals there, and transforming back after sampling.
The efficiency of this kind of continuous relaxation highly depends on the properties of the relaxed continuous distributions which may be arbitrarily difficult to sample from, such as being highly multi-modal. 
To avoid these pitfalls, for discrete distributions which can be displayed as continuous, differentiable functions accepting real-valued inputs but are evaluated only on a discrete subset of their domain, Gibbs-with-gradient proposal~\citep{grathwohl2021oops} and discrete Langevin proposal~\citep{zhang2022langevin} use gradients to inform discrete updates directly for these discrete distributions rather than transport the discrete domain to a continuous one.
However, most discrete distributions in the real world do not have a natural continuous extension, or the natural extension is still not differentiable.
This is why we propose Newton proposal.

\textbf{Locally-Balanced Proposal}
Based on local neighborhood information at the current location, the locally-balanced proposal~\citep{zanella2020informed} is an informed framework that is applicable to both discrete and continuous spaces.
When sampling from discrete distributions, it does not require natural differentiable extensions. 
Locally-balanced proposals have been extended to continuous-time Markov processes~\citep{power2019accelerated} and have been tuned via mutual information~\citep{sansone2022lsb}. 
It has also been used in Multiple-try Metropolis (MTM) algorithms to achieve fast convergence~\citep{gagnon2022improving}.
It is very expensive to construct locally-balanced proposals when the local neighborhood is large or the dimension is high, preventing them from making large moves in discrete spaces. 
The path auxiliary proposal~\citep{sun2021path} explores a larger neighborhood by making a sequence of small moves. 
An adaptive locally-balanced proposal~(ALBP)~\citep{sun2022optimal} has been proposed to determine the update size automatically.
However, it still only updates one coordinate per gradient computation and the update has to be done in sequence.
On the contrary, our Newton proposal can update many coordinates in parallel.

\section{PRELIMINARY}
We consider sampling from a target distribution
\begin{equation}
    \pi(\theta)=\frac{1}{Z} \exp (U(\theta)), \quad \forall \theta \in \Theta
\end{equation}
where $\theta$ is a $d$-dimensional variable, $\Theta$ is a finite variable domain, the energy function $U$ is a scalar-valued function, and $Z$ is the normalizing constant for $\pi$ to be a distribution. 
In this paper, we restrict $\Theta$ to a factorized domain, $i.e.$, $\Theta=\prod_{i=1}^{d} \Theta_{i}$, and mainly consider $\Theta$ to be $\{0,1\}^{d}$ or $\{0,1, \ldots, L-1\}^{d}$, which correspond to the binary variable and the categorical one, respectively. 

As we state in related work, Locally-balanced proposal~\citep{zanella2020informed} does not require a natural continuous distribution, so it is a flexible framework to build efficient and informed proposals for discrete distributions:
\begin{equation}\label{eq:Q0}
    Q_{g, \sigma}(\theta, d \theta^{\prime})\propto g\left(\frac{\pi(\theta^{\prime})}{\pi(\theta)}\right) K_\sigma(\theta, d \theta^{\prime}),
\end{equation}
where $g$ is a continuous function from $[0, \infty)$ to itself satisfying $g(t)=\operatorname{tg}(1 / t),~\forall t>0$.
$K_\sigma(\theta, d \theta^{\prime})$ is a symmetric kernel and $\sigma$ is a scale parameter.

When we set $g(t)=\sqrt{t}$, $K_\sigma(x, \cdot)=N\left(x, \sigma^2\right)$ and $\alpha=\sigma^2$ as the well-known Metropolis-Adjusted Langevin Algorithm~(MALA) proposal~\citep{roberts1998optimal}, we get an informed proposal as
\begin{equation}\label{eq:Q}
    q_0\left(\theta^{\prime} \mid \theta\right)\propto\exp\left(\frac{U(\theta^{\prime})-U(\theta)}{2}-\frac{\|\theta^{\prime}-\theta\|^2}{2\alpha}\right),
\end{equation}
where the local difference $U(\theta^{\prime})-U(\theta)$ shows the likelihood ratio between a given input $\theta$ and other discrete states $\theta^{\prime}$.
In case where summing over the full space of $\theta^{\prime}$ is so expensive that the normalizing constant becomes intractable, the locally-balanced proposal often restricts its domain to a small neighborhood.
For example, the Gibbs-with-gradient proposal~\citep{grathwohl2021oops} only considers local moves inside a Hamming ball with small window sizes.

\textbf{Finite Difference}
Finite Difference is a mathematical expression of the form $f(x+b)-f(x+a)$, which is an approximation of derivatives.
Specifically, a forward finite difference, denoted $\Delta_h[f]$, of a function $f$ is defined as
$$
\Delta_h[f](x)=f(x+h)-f(x).
$$
When the window size $h=1$, $h$ can be omitted:
$$
\Delta[f](x)=f(x+1)-f(x).
$$

As for the finite difference with respect to a vector, let's consider $x \in \{0,1, \ldots, L-1\}^{d}$ as a $d$-dimensional vector and $f: \{0,1, \ldots, L-1\}^{d} \rightarrow \mathbb{R}$ as a scalar-valued function. 
The finite difference of $f$ with respect to the vector $x$ is defined as
$$
\Delta[f](x)=\left(\Delta[f](x)_1,\ldots, \Delta[f](x)_d\right).
$$
Specifically,
$$
\Delta[f](x)_i=f\left(\neg_i x\right)-f(x),
~\forall i\in \{1, \ldots, d\},
$$ 
where $\neg_i x$ changes the $i$-th coordinate from $x_i$ to $x_i+1$ and keeps the other $d-1$ coordinates the same as $x$.

\textbf{Newton's Series Expansion}
As the discrete analog of the continuous Taylor expansion, Newton's series expansion is used to approximate discrete functions.
In Newton's series expansion, we use finite differences instead of gradients to indicate neighborhood information.

Let's consider the scalar version first.
The Newton series consists of the terms of the Newton forward difference equation:
$$
f(x)=\sum_{k=0}^{\infty} \frac{\Delta^k[f](a)}{k !}(x-a)_k
$$
where $(x)_k=x(x-1)(x-2) \cdots(x-k+1)$ and $\Delta^k[f](x)$ represents $k$-th order forward finite difference defined as
$$
\Delta^k[f](x)=\sum_{i=0}^k\left(\begin{array}{l}k \\ i\end{array}\right)(-1)^{k-i} f(x+i).
$$
Specifically, the first-order Newton's series expansion is
\begin{equation}\label{eq:1st_sim}
    f(x)\approx f(a)+\Delta[f](a)(x-a).
\end{equation}

When $x$ and $a$ are $d$-dimensional vectors, $\Delta[f](a)$ is also a $d$-dimensional vector and the corresponding first-order Newton's series expansion becomes
\begin{equation}\label{eq:1st_sim_vec}
    f(x)\approx f(a)+\Delta[f](a)^{\top}\cdot (x-a).
\end{equation}
In this paper, we use the first-order Newton's series expansion under $h=1$ to approximate the likelihood of discrete moves.

\section{EFFICIENT INFORMED PROPOSALS VIA NEWTON'S SERIES APPROXIMATION}
To sample from discrete distributions efficiently, we propose Newton proposal.
We use Newton's series expansion to approximate the likelihood of making discrete updates, and factorize the discrete domain coordinatewise to reduce the computation cost significantly.

\subsection{Informed Proposal via Newton's Series Approximation}
Consider a common $d$-dimensional case $\Theta=\{0,1,\cdots,L-1\}^{d}$.
In each iteration, several coordinates are flipped and we update the current samples $\theta$ to $\theta^{\prime}$.
We use a first-order forward Newton's series expansion with window size $h=1$ to approximate $U(\theta^{\prime})-U(\theta)$:
\begin{equation}\label{eq:1st}
    U(\theta^{\prime})-U(\theta)\approx{\Delta[U](\theta)}^{\top} \cdot(\theta^{\prime}-\theta).
\end{equation}

We use the Newton's series expansion in \eqref{eq:1st} to approximate the local difference $U(\theta^{\prime})-U(\theta)$ in \eqref{eq:Q}:
\begin{equation}\label{eq:square}
\begin{aligned}
    &q\left(\theta^{\prime} \mid \theta\right)=\widehat{q_0}\left(\theta^{\prime} \mid \theta\right)\\
    =&\frac{1}{Z_{\Theta}(\theta)}\exp\left(\frac{{\Delta[U](\theta)}^{\top}\cdot(\theta^{\prime}-\theta)}{2}-\frac{\|\theta^{\prime}-\theta\|^2}{2\alpha}\right)\\
    \propto &\exp{\Bigg(\frac{1}{2\alpha}\left(-\big(\theta^{\prime}-\theta\big)^2+\alpha\Delta[U](\theta)^{\top}\cdot(\theta^{\prime}-\theta)\right)\Bigg)}\\
    &\cdot \exp{\Big(-\frac{\alpha}{8}\Delta[U](\theta)^2\Big)}\\
    =&\exp{\left(-\frac{1}{2\alpha}\|\theta^{\prime}-\theta-\frac{\alpha}{2}\Delta[U](\theta)\|^2\right)}.
\end{aligned}
\end{equation}
We add a term in the fourth line to get the perfect square form because $\Delta[U](\theta)$ is independent of $\theta^{\prime}$ and will not affect the normalized result (see the appendix for detailed proof).

In this way, we obtain the new proposal in a perfect square form by Newton's series expansion:

\begin{equation}\label{eq:gauss}
q\left(\theta^{\prime} \mid \theta\right)= \frac{\exp \left(-\frac{1}{2 \alpha}\left\|\theta^{\prime}-\theta-\frac{\alpha}{2} \Delta[U](\theta)\right\|_2^2\right)}{Z_{\Theta}(\theta)}
\end{equation}
where the normalizing constant is summed over $\Theta$:
\begin{equation}\label{eq:normalizing}
    Z_{\Theta}(\theta)=\sum_{\theta^{\prime} \in \Theta} \exp\left(-\frac{\left\|\theta^{\prime}- \theta - \frac{\alpha}{2} \Delta[U](\theta)\right\|_2^2}{2 \alpha}\right).\nonumber
\end{equation}
In a word, finite difference in Newton's series approximation serves as a guide when exploring the discrete space, similar to what gradients do in proposals for continuous distributions.
It provides neighborhood information about the target distribution so that the sampler can propose new states more informatively rather than ``blindly''.
Moreover, the perfect square form gives us possibility to accelerate the computation without restrict our domain in a small neighborhood, which we will discuss in detail later.

\subsection{Efficient Newton Proposal via Coordinatewise Factorization}
The computation cost of the informed proposal in~\eqref{eq:gauss} depends on the normalizing constant $Z_{\Theta}(\theta)$ in \eqref{eq:normalizing}, which needs to sum up all states in the discrete space.
Therefore, it is desirable to narrow down the space to make $Z_{\Theta}(\theta)$ tractable.

Unlike most locally-balanced proposals which restrict the domain to a small neighborhood, $e.g.$, a hamming ball~\citep{zanella2020informed,grathwohl2021oops}, a key feature of the proposal \eqref{eq:gauss} is that it is displayed as a Euclidean norm and can be factorized coordinatewise~\citep{zhang2022langevin}. 
To see this, we write \eqref{eq:gauss} as $q\left(\theta^{\prime} \mid \theta\right)=\prod_{i=1}^{d} q_{i}\left(\theta_{i}^{\prime} \mid \theta\right)$
where $q_{i}\left(\theta_{i}^{\prime} \mid \theta\right)$ is a simple categorical distribution:
\begin{equation}\label{eq:newton}
    \operatorname{Cat} \left(\sigma\left(\frac{1}{2}\Delta[U](\theta)_i\left(\theta_{i}^{\prime}-\theta_{i}\right)-\frac{\left(\theta_{i}^{\prime}-\theta_{i}\right)^{2}}{2 \alpha}\right)\right).
\end{equation}
Here, $\operatorname{Cat}$ stands for categorical distribution, $\sigma$ denotes SoftMax function and $\theta_{i}^{\prime} \in \Theta_{i}$. 
Recall that $\Delta[U](\theta)_i=U\left(\neg_i \theta\right)-U(\theta),~\forall i\in \{1, \ldots, d\}$
where $\neg_i \theta$ changes the $i$-th coordinate from $\theta_i$ to $\theta_i+1$ while keeping the other coordinates the same as $\theta$.

Since both the domain $\Theta$ and the proposal over all coordinates in~\eqref{eq:gauss} can be factorized coordinatewisely, we can update each coordinate in parallel, thus greatly speeding up the computation. 
In this way, Newton proposal fills in the blank in bottom right of Table \ref{table:compare}.
It enables us to sample from high-dimensional complex discrete distributions with better mixing and faster convergence, no matter whether the discrete distribution has an natural differentiable extension.

When modeling the proposal distribution over all coordinates jointly, the overall cost of constructing the proposal in \eqref{eq:gauss} is $\mathcal{O}\left(L^d\right)$ for $\{0,1, \ldots, L-1\}^{d}$.
Thanks to the coordinatewise factorization, the cost of Newton proposal is reduced to $\mathcal{O}(Ld)$. 
This allows the sampler to explore the full space with the neighborhood information without paying a prohibitive cost.

\subsection{A Variant: With a MH Correction}
It is optional to add a Metropolis-Hastings (MH) step~\citep{metropolis1953equation, zhang2022langevin}, which is usually combined with proposals to make the Markov chain reversible.
Specifically, after generating the next position $\theta^{\prime}$ from a distribution $q(\cdot \mid \theta)$, the MH step accepts it with probability
\begin{equation}\label{eq:acc}
    \min \left(1, \exp \left(U\left(\theta^{\prime}\right)-U(\theta)\right) \frac{q\left(\theta \mid \theta^{\prime}\right)}{q\left(\theta^{\prime} \mid \theta\right)}\right).
\end{equation}
By rejecting some of the proposed states, the Markov chain is guaranteed to converge asymptotically to the target distribution.
The sampler with our Newton proposal is outlined in Algorithm \ref{alg:DPULA}.

We call Newton proposal without the MH step as unadjusted Newton algorithm~(\textbf{UNA}) and that with the MH step as Metropolis-adjusted Newton algorithm~(\textbf{MANA}).
Similar to MALA and ULA in continuous spaces~\citep{grenander1994representations,roberts2002langevin}, MANA contains $2Ld$ (for finite difference computation) plus $2$ (for the MH correction) function evaluations and is guaranteed to converge to the target distribution.
Although UNA may have asymptotic bias, it only requires $Ld$ function evaluations, which is valuable especially when performing the function evaluation is expensive such as in large-scale Bayesian inference~\citep{welling2011bayesian,durmus2019high}.

\begin{algorithm}
\caption{Samplers with Newton Proposal.}
\label{alg:DPULA}
\begin{algorithmic}[0]
\STATE \textbf{given:} Stepsize $\alpha$.
\STATE \textbf{loop}
\begin{ALC@g}
\FOR{$i=1,\dots,d$}
\STATE \textcolor{blue}{(Can be done in parallel)}
\STATE \textbf{construct} $q_i(\cdot\mid \theta)$ as in Equation \eqref{eq:newton}
\STATE \textbf{sample} $\theta_{i}^{\prime} \sim q_{i}(\cdot \mid \theta)$
\ENDFOR
\STATE $\triangleright$ Optionally, do the MH step
\STATE \textbf{compute} $q\left(\theta^{\prime} \mid \theta\right)=\prod_{i} q_{i}\left(\theta_{i}^{\prime} \mid \theta\right)$
\STATE \qquad ~and $q\left(\theta \mid \theta^{\prime}\right)=\prod_{i} q_{i}\left(\theta_{i} \mid \theta^{\prime}\right)$
\STATE \textbf{set} $\theta \leftarrow \theta^{\prime}$ with probability in Equation \eqref{eq:acc}
\end{ALC@g}
\STATE \textbf{end loop}
\STATE \textbf{output:} samples $\left\{\theta_k\right\}$.
\end{algorithmic}
\end{algorithm}

\section{AN ALTERNATIVE VIEW OF NEWTON PROPOSAL}
After getting the Newton proposal via Newton's series, we give an alternative way to derive Newton proposal via multilinear extension, which gives us more intuition about the efficient informed proposal.
From the multilinear extension viewpoint, we find further connection with Discrete Langevin Proposal~(DLP)~\citep{zhang2022langevin}.

\subsection{An Equivalent Form via Multilinear Extension}\label{sub:multilinear}
In addition to approximating the likelihood ratio of flipping each dimension with Newton's series expansion, we find that our Newton proposal can also be obtained by conducting Taylor expansion on the multilinear extension of the discrete distribution.
This connection gives us another interesting viewpoint of the Newton proposal.
We briefly show the equivalence between these two viewpoints in the binary case here and put the categorical case and detailed proof in the appendix.

Let us consider the $d$-dimensional binary distribution.
The coordinates can be denoted as a finite set $D=\left\{1,\cdots,d\right\}$. 
The sampling process corresponds to choosing which coordinate to flip, so the discrete distribution is a set function defined over the power set of $D$ as $f: 2^{D} \rightarrow \mathbb{R}$.
A discrete distribution may not have a natural continuous extension, but its multilinear extension $F:[0,1]^{d} \rightarrow \mathbb{R}$ can always be defined as :
\begin{equation}\label{eq:multilinear}
    F({\theta})=\sum_{S \subseteq D} f(S) \prod_{i \in S} \theta_{i} \prod_{i \in D \backslash S}\left(1-\theta_{i}\right).
\end{equation}

As we can see, $F(\theta)$ is a continuous, differentiable function which accepts real-valued inputs from the interval $[0,1]^d$.
This makes it possible to approximate the likelihood of discrete moves with Taylor series expansion.
Besides, an inspiring fact is that $F(\theta)$ keeps the same value with $f$ when they are evaluated on the discrete subset $\{0,1\}^d$.
For $i \in D$, since $F$ is linear in $\theta_{i}$, we have the partial derivative of $F({\theta})$ as:
\begin{eqnarray}\label{eq:1der}
\begin{aligned}
\frac{\partial F}{\partial \theta_{i}}({\theta})=&F\left(\theta_{1}, \ldots, \theta_{i-1}, 1, \theta_{i+1}, \ldots, \theta_{d}\right)\\
-&F\left(\theta_{1}, \ldots, \theta_{i-1}, 0, \theta_{i+1}, \ldots, \theta_{d}\right)\nonumber
\end{aligned}
\end{eqnarray}
We can see that the partial derivative measures the difference between the energy function of the original state and the flipped one, which is exactly the finite difference of the two states.
Since $F$ and $f$ take the same value on the discrete domain, the Newton proposal can be equivalently obtained by conducting Taylor expansion on the multilinear extension of the target discrete distribution.

As for categorical distribution, we need to decide not only which coordinate to flip, but also which level to flip to.
Fortunately, the differentiable function $F(\theta)$ can be obtained with a generalized multilinear extension~\citep{sahin2020sets}, on which the likelihood ratio of flipping each coordinate to any level can be defined.
The detailed algorithm is in the appendix.

\subsection{Comparison with Discrete Langevin Proposal}
Our Newton proposal and the Discrete Langevin Proposal~(DLP)~\citep{zhang2022langevin} can be seen as two different approximations of the locally-balanced proposal~\citep{zanella2020informed} when taking functions $g$ and $K$ like MALA, as shown in \eqref{eq:Q}.

DLP is motivated by utilizing gradients to guide the sampling and inference. 
Thus it requires that the discrete distribution can be displayed as a differentiable function which is only evaluated on a discrete domain, $e.g.$, Ising models, so that the gradient can be defined.
In this way, DLP can be viewed as a first-order Taylor series approximation to the local difference term inside $q_0\left(\theta^{\prime} \mid \theta\right)$ in \eqref{eq:Q} with:
$$
U(x)-U(\theta) \approx \nabla U(\theta)^{\top}(x-\theta),~\forall x \in \Theta.
$$
In contrast, our Newton proposal circumvents this shortcoming by using Newton's series expansion, a natural tool to approximate discrete functions.
Specifically, Newton proposal approximates the local difference in $q_0\left(\theta^{\prime} \mid \theta\right)$ with:
$$
U(x)-U(\theta) \approx \Delta [U](\theta)^{\top} (x-\theta),~\forall x \in \Theta.
$$
The finite difference $\Delta [U](\theta)$ is defined on the grid-like discrete domain and can also guide to explore the discrete space like what gradients do in continuous relaxation-based methods.

In addition to the differences in methods and requirements, our Newton proposal has more general applications than DLP:
(1) When the discrete distribution has a natural differential extension, such as Ising model, Restricted Boltzmann Machines~(RBMs) and Potts model,
DLP and the Newton proposal both work.
In some special cases such as some Ising models, when the multilinear extension and the natural continuous extension of the original discrete distribution are the same, DLP and Newton proposal have the same results.
(2) The strict requirement about natural differential relaxation limits DLP to be widely used in more complex scenarios.
When the target discrete distribution lacks a differentiable extension, DLP does not work while
the Newton proposal is still well-suited for these tasks, such as the facility location problem, text summarization, etc.

For the computation cost, DLP contains one gradient computation whose cost depends on $d$ and $L$, while the Newton proposal contains $Ld$ function evaluations. Both gradient and function computations can be done in parallel.

\section{THEORETICAL ANALYSIS}
In this section, we analyze the asymptotic convergence of UNA and the asymptotic efficiency of MANA. 
Specifically, we first prove in Section \ref{sec:una} that when the stepsize $\alpha \rightarrow 0$, the asymptotic bias of UNA is zero when the discrete distribution is a second-order modular function, whose gain reduction keeps the same for any set~\citep{korula2018online}~(see the rigorous definition in the appendix).
Later in Section \ref{sec:mana}, we derive the asymptotic efficiency of MANA.
\subsection{Asymptotic Convergence of UNA for Second-order Modular Functions}\label{sec:una}
Besides the property of the proposal itself, the effectiveness of a proposal also depends on how close its underlying stationary distribution is to the target distribution because if it is far, even if using the MH step to correct the bias, the acceptance probability will be very low. 
We consider a second-order modular distribution, which appears in common tasks such as Ising models. 
The following theorem summarizes UNA’s asymptotic accuracy for such discrete distributions.

\textbf{Theorem 1.} When the discrete distribution is a second-order modular function~\citep{korula2018online}, 
its multilinear extension $F({\theta})$ is quadratic which can be expressed as $F({\theta})=\theta^{\top}A\theta+b^{\top}\theta$.
The Markov chain following transition $q(\cdot \mid {\theta})$ in \eqref{eq:newton} (i.e. UNA) is reversible with respect to some distribution $\pi_{\alpha}$ and $\pi_{\alpha}$ converges weakly to $\pi$ as $\alpha \rightarrow 0$. 
In particular, let $\lambda_{\min }$ be the smallest eigenvalue of $A$, then for any $\alpha>0$,
$$
\left\|\pi_{\alpha}-\pi\right\|_{1} \leq Z \cdot \exp \left(-\frac{1}{2 \alpha}-\frac{\lambda_{\min }}{2}\right).
$$

Theorem 1 shows that the asymptotic bias of UNA decreases at a $\mathcal{O}(\exp (-1 /(2 \alpha))$ rate which vanishes to zero as the stepsize $\alpha \rightarrow 0$.
Besides, the asymptotic bias of UNA is related to the smallest eigenvalue of $A$.

\begin{figure}
    \centering
    \includegraphics[height=3.5cm]{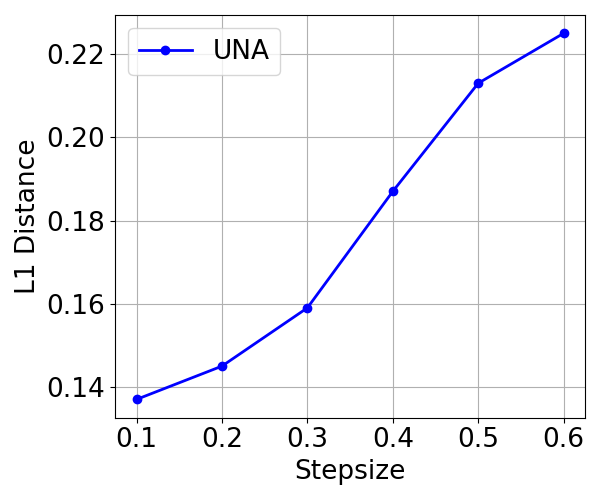}
    \caption{UNA with varying stepsizes on an Ising model.}
    \label{fig:verify1-1}
\end{figure}

\textbf{Example.} Let's consider the well-known model in thermodynamic systems, the Ising model.
Note that its distribution is second-order modular:
\[
f(D) = \sum_{u,v\in D}A(u,v)+\sum_{u\in D} b(u)
\]
where $A(u,v)$ represents the interaction between any two adjacent sites $u,v\in D$, and a site $u \in D$ has an external magnetic field $b(u)$ interacting with it.
We run UNA with varying stepsizes on a 2 by 2 Ising model, as shown in Figure \ref{fig:verify1-1}. 
For each stepsize, we run the chain long enough to ensure its convergence. 
The results clearly show that the distance between the stationary distribution of UNA and the target distribution decreases as the stepsize decreases. 

\subsection{Asymptotic Efficiency of MANA}\label{sec:mana}
To understand the asymptotic efficiency of MCMC transition kernels, we study the asymptotic variance and the spectral gap of the kernel. The asymptotic variance is defined as
$$
\operatorname{var}_p(h, Q)=\lim _{T \rightarrow \infty} \frac{1}{T} \operatorname{var}\left(\sum_{t=1}^T h\left(x_t\right)\right)
$$
where $h: \mathcal{X} \rightarrow R$ is a scalar-valued function, $Q$ is a $p$-stationary Markov transition kernel. 
The asymptotic variances measures the additional variance incurred when using sequential samples from $Q$ to estimate $E_p[h(x)]$. 
The spectral gap is defined as
$$
\operatorname{Gap}(Q)=1-\lambda_2
$$
where $\lambda_2$ is the second largest eigenvalue of the transition probability matrix of $Q$. 
For transition probability matrices with non-negative eigenvalues, the spectral gap is related to the mixing time, with larger values corresponding to faster mixing~\citep{levin2017markov}.

Since our method approximates $Q_{g, \sigma}(\theta, d \theta^{\prime})$ in \eqref{eq:Q0}, we should expect some decrease in efficiency.
We characterize this decrease in terms of the asymptotic variance and the spectral gap, under the Lipschitz-like assumption on $\Delta[U](\theta)$.
In particular, we show that the decrease is a constant factor that depends on the Lipschitz constant of $\Delta[U](\theta)$ and the dimension of the target distribution.

\textbf{Theorem 2}
Let $Q\left(\theta^{\prime}, \theta\right)$ and $\tilde{Q}\left(\theta^{\prime}, \theta\right)$ be the Markov transition kernels given by the Metropolis-Hastings algorithm using the locally-balanced proposal $q_0\left(\theta^{\prime} \mid \theta\right)$ and our approximation $q\left(\theta^{\prime} \mid \theta\right)$. 
Let the finite difference $\Delta[U](\theta)$ has an Lipschitz-like property with constant $L$, and $\pi(\theta)=\frac{\exp (U(\theta))}{Z}$. 
Then it holds
\begin{enumerate}
    \item $\operatorname{var}_{\pi}\left(h, \tilde{Q}\right) \leq \frac{\operatorname{var}_{\pi}(h, Q)}{c}+\frac{1-c}{c} \cdot \operatorname{var}_{\pi}(h)$.
    \item $\operatorname{Gap}\left(\tilde{Q}\right) \geq c \cdot \operatorname{Gap}(Q)$ 
\end{enumerate}
where $c=e^{-\frac{1}{2} L D^2}$ and $D=\sup _{\theta^{\prime} \in \Theta}\left\|\theta^{\prime}-\theta\right\|$.

\textbf{Remark.} We can see that the constant $D$ is correlated with the dimension of the target discrete distribution.
In high-dimensional scenarios, $D$ will be a large constant, leading to loose bounds of the asymptotic variance and spectral gap.
However, on one hand, there is a gap between theory and experiment~\citep{kwisthout2013bridging}.
It may be hard to achieve the bound in practice.
On the other hand, we can add a slight restriction on the number of changed coordinates in a single step.
In this way, $D$ can be reduced to a small number as we expect, and we can get tighter bounds in theory.

\section{EXPERIMENTS}
We conduct a comprehensive empirical evaluation for Newton proposal on synthetic and real-world sampling tasks.
The unadjusted and Metropolis-adjusted Newton proposals are denoted as UNA and MANA, respectively.
We release the code at \href{https://github.com/DongyaoZhu/Newton-Proposal-for-Discrete-Sampling}{https://github.com/DongyaoZhu/Newton-Proposal-for-Discrete-Sampling}.
Baselines and evaluation tasks are described below.

\textbf{Baselines.} We compare the performance of Newton proposals with widely-used sampling methods for discrete distributions, including (1) two Gibbs-based methods: Gibbs sampling, Gibbs with Gradient~(GWG)~\citep{grathwohl2021oops}; (2) two methods which perform sampling in a continuous
space by gradient-based methods and then transforms the collected samples to the original discrete space: discrete Stein Variational Gradient Descent (D-SVGD)~\citep{han2020stein} and relaxed MALA (R-MALA)~\citep{grathwohl2021oops}; (3) one gradient-based method which requires the target discrete distribution to have a differential relaxation: Discrete Langevin Proposal~(DLP)~\citep{zhang2022langevin} and (4) the locally-balanced sampler (LB)~\citep{zanella2020informed}.
Specifically, we denote DULA and DMALA for unadjusted and Metropolis-adjusted DLPs, respectively.
All methods are implemented in Pytorch and we use the official release of code from previous papers when possible.

\textbf{Evaluation Tasks.}
(1) Discrete distributions without natural differentiable extensions. Since GWG and DLP require gradients and can not be applied, we mainly compare the Newton proposal with Gibbs and LB in these tasks.
(2) Discrete distributions with natural differentiable extensions.
We also apply Newton proposal to these distributions such as the Ising model which is binary and Potts model which is categorical, to show the broad applicability of our method. 
In this scenario, we also compare Newton proposal with continuous relaxation methods (GWG and DLP) and the results are included in the appendix.

\subsection{Facility Location}
In the facility location task, we are given a set of facilities denoted $\mathcal{V}$ and a set of $m$ customers to decide whether to open a facility or not~\citep{krause2008efficient}, corresponding to sampling from a binary distribution.
If the $i$-th facility $(i\in \mathcal{V})$ is opened, then it provides service of value $c_{i, j}$ to customer $j~(j\in \left\{1,\cdots,m\right\})$.
We suppose that each customer chooses the opened facility with highest value, then the total value provided to all customers is $\sum_{j=1}^m \max _{i \in \mathcal{V}} c_{i, j}$. 
Besides, we penalize the
number of selected facilities to ensure the most total utility with a small number of facilities.
Therefore, the distribution of the facility location model can be represented as
$
f(S)=\sum_{j=1}^{m} \max _{i \in \mathcal{V}} c_{i, j}-\lambda|\mathcal{V}|,
$
where $\lambda$ is a hyperparameter, controlling the strength of the penalty term.
To evaluate the Newton proposal on facility location task, we generate the utility matrix $\bm{C}$ from a gaussian mixture model with $m=64$ and $|\mathcal{V}|=15$.
We run 30000 iterations and set the stepsizes of MANA and UNA as 1 and 0.2, respectively.

\begin{figure}[t]
    \centering
    \subfigure[$\log{\text{RMSE}}$ w.r.t. Iterations]{
        \includegraphics[height=3.7cm]{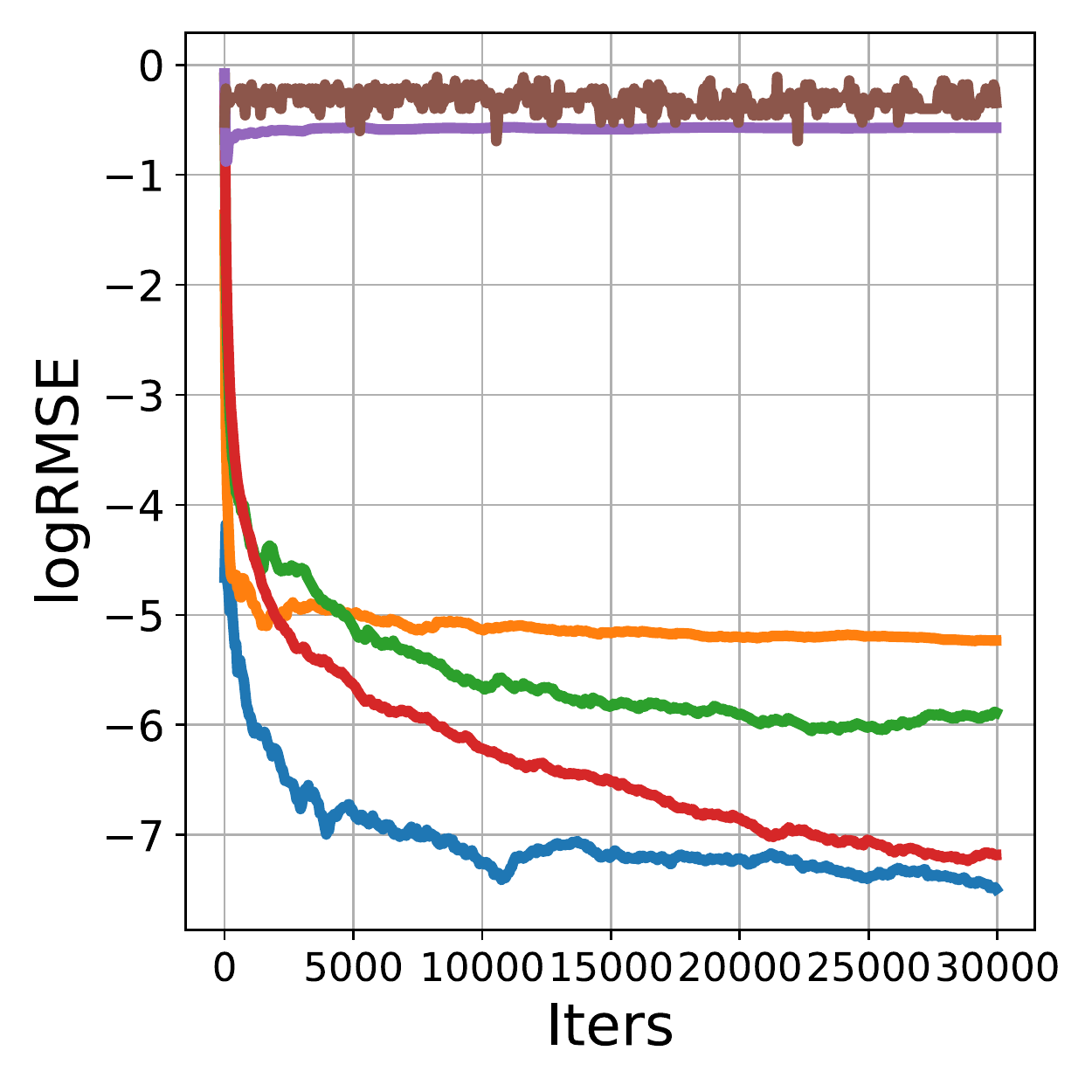}\label{fig:location_iter}
    }
    \subfigure[$\log{\text{RMSE}}$ w.r.t. Runtime]{
        \includegraphics[height=3.7cm]{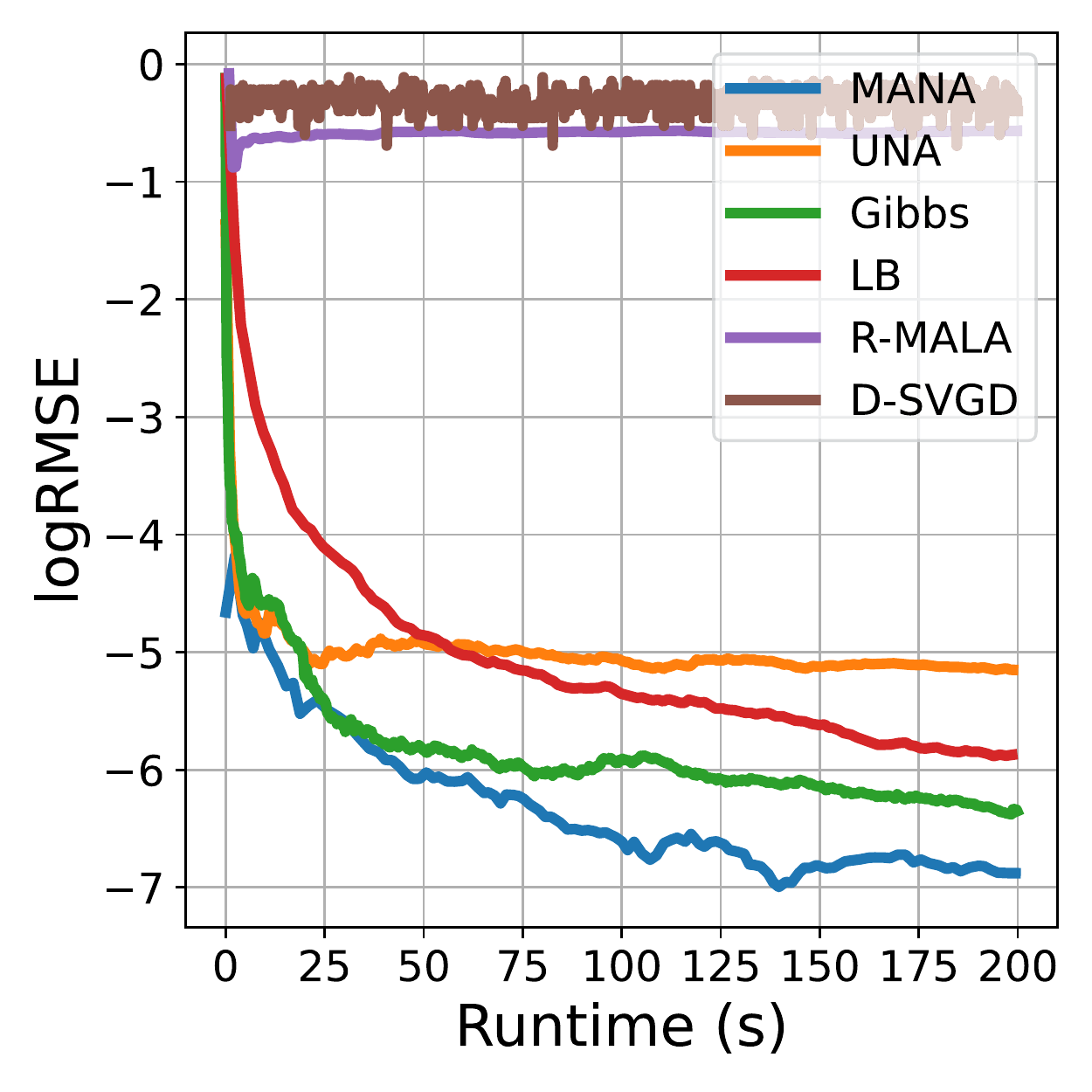}\label{fig:location_time}
    }
    \caption{Facility location model results. MANA outperforms the baselines in both number of iterations and running time.}
    \label{fig:location}
\end{figure}

\begin{figure}[t]
    \centering
    \subfigure[AccRate, \#Changed Dims w.r.t. Stepsize]{
        \includegraphics[height=3.4cm]{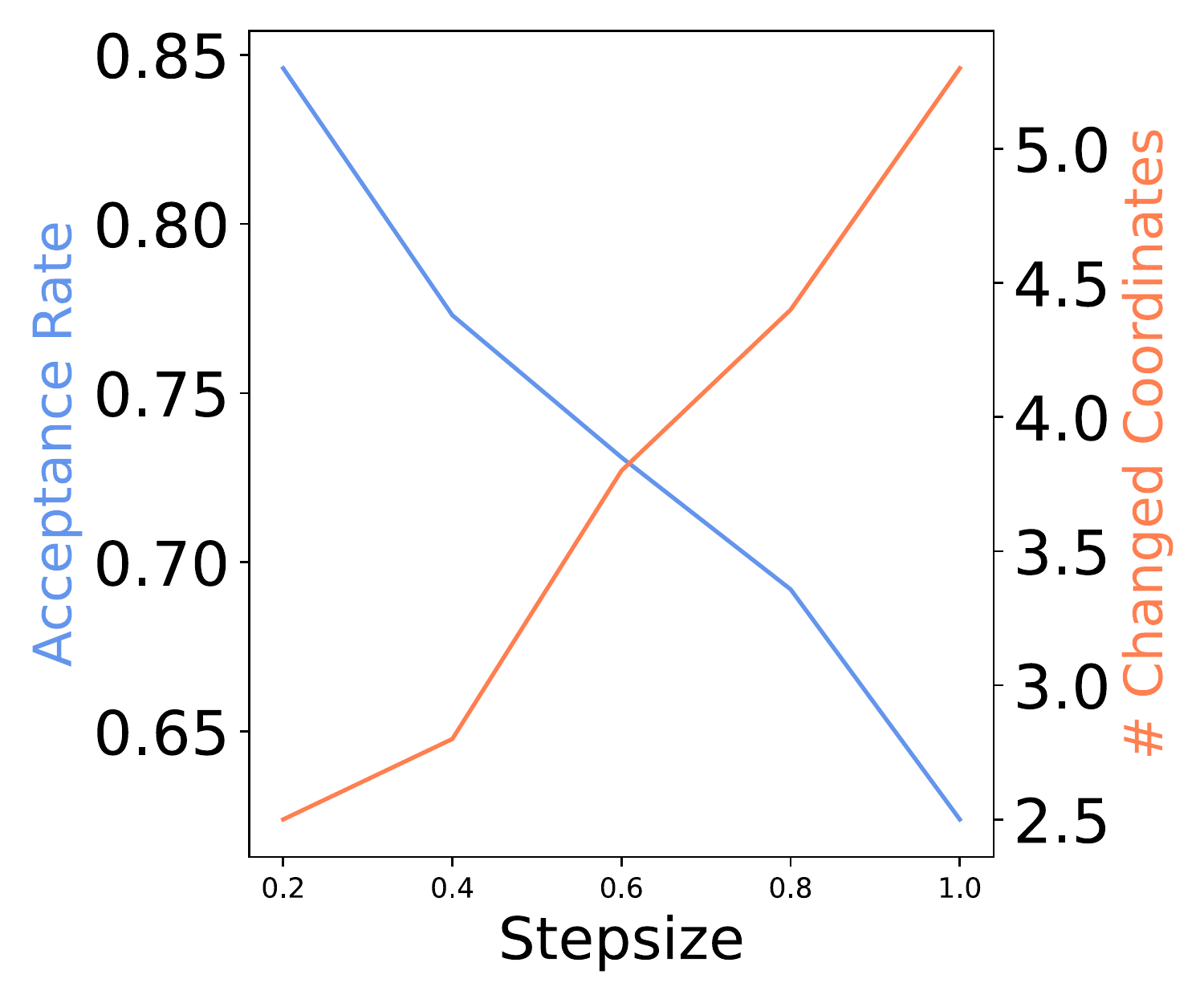}\label{fig:ncor}
    }
    \subfigure[ESS of Proposals]{
        \includegraphics[height=3.4cm]{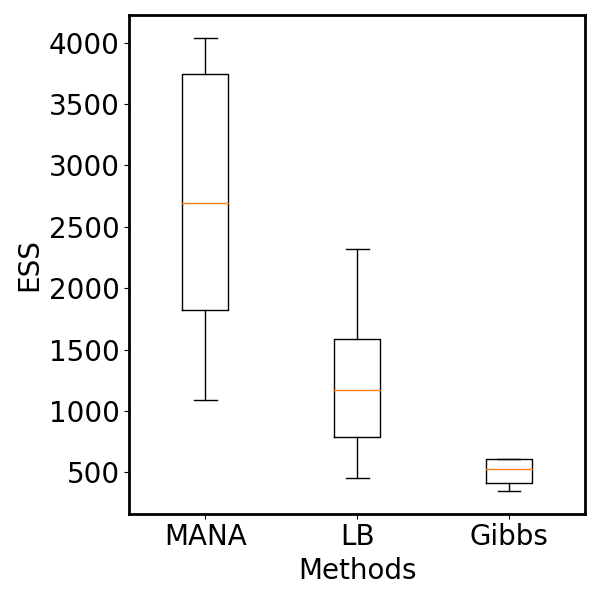}\label{fig:loc_ess}
    }
    \caption{Facility location sampling results. \\
    \textbf{Left:} Newton proposal keeps a higher acceptance rate. 
    \textbf{Right:} MANA yields the largest effective sample size (ESS) among all the methods compared.}
    \label{fig:location2}
\end{figure}

We first compare the root-mean-square error (RMSE) between the estimated mean and the true mean under $\lambda=10$ in Figure \ref{fig:location}.
The blue lines of MANA are both below other lines, indicating that MANA is the fastest to converge in terms of both iterations and running time.
This demonstrates (1) sampling in the original discrete
space is important: D-SVGD and R-MALA get poor results because this task is complex and the relaxed distributions are hard to sample from; (2) the finite difference makes the exploration over the discrete space more informative rather than ``blind'' compared to Gibbs; (3) the MH step enables MANA to make larger and more effective moves with a larger step size without worrying about unconvergence compared to UNA; (4) changing many coordinates in one step accelerates the convergence compared to LB and Gibbs.
We then show that Newton proposal can change multiple coordinates in one iteration while still maintaining a high acceptance rate in Figure \ref{fig:ncor}. 
When the stepsize $\alpha=1$, on average MANA can change \textbf{5.3} coordinates in one iteration with an acceptance rate \textbf{$62.4 \%$} in the MH step, while the acceptance rate of LB proposal is only $35\%$. 
In Figure \ref{fig:loc_ess}, we compare the effective sample size (ESS) for exact samplers ($i.e.$, having the target distribution as its stationary distribution).
MANA outperforms other methods, indicating the correlation among its samples is low due to making significant updates in each step.

\begin{table*}[t]
\centering
\caption{Performance of sampling methods on Extractive Text Summarization on DUC-2002 dataset. We report $\mathcal{F}(S)$ at 500, 750 and 1000 steps, as well as ESS, runtime of 1000 steps~(s), the ROUGE-2 recall~(R), F-measure~(F) and Precision~(P)~(\%).}
\begin{small}
\begin{tabular}{l|cccccccc}
\hline\hline
Method & $\mathcal{F}(500)$        & $\mathcal{F}(750)$        & $\mathcal{F}(1000)$       & R             & F             & P  & ESS & Runtime          \\ \hline
\textbf{MANA}   & \textbf{6.28} & \textbf{6.40} & \textbf{6.46} & \textbf{8.72} & \textbf{8.85} & \textbf{10.96} & \textbf{57.2} & 6.7\\
LB     & 6.28          & 6.39          & 6.42          & 7.91   & 8.40   & 10.68  & 41.8 &  9.9\\
Gibbs  & 6.01          & 6.12          & 6.42          & 7.91   & 8.31    & 10.62   &15.2 &\textbf{1.0}\\
\hline\hline
\end{tabular}
\label{tab:text}
\end{small}
\end{table*}

\subsection{Extractive Text Summarization}
Extractive text summaries are formed by selecting several sentences $S$ from source documents that best fit certain quality measurements. ~\citep{lin2011class} designed their metrics on $S$ for both similarity and diversity, formally defined as $\mathcal{F}(S)=\mathcal{L}(S)+\lambda \mathcal{R}(S)$ subject to some cost constraint $C(S) \leq K$, 
where $\mathcal{L}(S)$ measures the coverage or "fidelity" of summary set $S$ to the document, $\mathcal{R}(S)$ rewards diversity in $S$, and $\lambda>0$ is a trade-off coefficient.
The undifferentiable distribution of the summary $f(S)=\max \mathcal{F}(S)$ makes the task well-suited for our methods based on pseudo-gradients. Futhermore, due to the limited time constraint, the proposals with non-parallel updates cannot explore enough into the distribution, while our Newton proposal will be able to avoid this issue with multidimensional updates.
We use an exponential decay schedule on step size to encourage faster convergence under limited time constraint. The final result is then given by a sample-wise majority vote algorithm ~\citep{wang2022self} on the collection of samples we produce.

We report results of MANA, LB and Gibbs sampler on DUC 2002 dataset~\citep{over2002introduction}, which contains about 30 sentences per document.
ROUGE scores ~\citep{lin2004rouge} are widely used for text summarization evaluation, and we compare ROUGE-2 scores (precision $P$, recall $R$, and F-measure $F$) of different samplers. 
In addition, we show the average objective scores $\mathcal{F}(S)$ at 500th, 750th and 1000th iteration, respectively. 
As shown in Table \ref{tab:text}, Newton proposal constantly outputs highest $\mathcal{F}(S)$ and ROUGE-2 scores under limited time constraints.

\begin{table}[t]
\centering
\caption{Performance of sampling methods on Image Retrieval on Holidays dataset. $\mathcal{F}(S)$ and mean Average Precision~(mAP) are reported at 500, 750, 1000 steps.}
\begin{small}
\begin{tabular}{l|cccc}
\hline\hline
Method & $\mathcal{F}(500$) & $\mathcal{F}(750)$ & $\mathcal{F}(1000)$ & mAP  \\ \hline
\textbf{MANA}   & \textbf{11.36}     & \textbf{11.37}     & \textbf{11.38}      & \textbf{0.84} \\
LB     & 11.04     & 11.05     & 11.05      & 0.55 \\
Gibbs  & 11.01     & 11.01     & 11.03      & 0.53\\
\hline\hline
\end{tabular}
\end{small}
\label{tab:IR}
\end{table}

\subsection{Image Retrieval}
Given a database of images and a query image, we look for a subset from the database that best matches the query. We follow the same settings in the extractive text summarization experiments, and we use the discontinuous score function $\mathcal{F}(S)$ proposed by ~\citep{yang2014submodular} to measure the matching of a particular collection of images to a query image. We empirically found that sample-wise majority vote algorithm ~\citep{wang2022self} did not perform well, thus we also propose a dimension-wise majority vote algorithm: given a collection of $N$ samples $X\in \{0, 1\}^{N, D}$, each dimension $d$ will have a count of $x_{d} = 1$, and the dimensions of top counts are selected (details in appendix).

We evaluate various methods on the INRIA Holidays Dataset \citep{jegou2008hamming} which consists of 1491 images (dimensions) and 500 queries. Our results in $\mathcal{F}(S)$ values and mean Average Precision~(mAP) are reported in Table \ref{tab:IR}. The $\mathcal{F}(S)$ shows that the sampler with Newton proposal quickly reaches and stably keeps a better solution to the maximization problem than other methods despite limited time constraint. Our high mean Average Precision demonstrates that our solution is a better approximation to the ground truth labels compared to other methods.

\section{CONCLUSION}
We propose a new gradient-like efficient informed proposal, the Newton proposal, for general discrete distributions.
This proposal better explores discrete spaces under the guidance of the finite difference produced by Newton's series expansion, which does not require natural differentiable expansions. Additionally, the factorization on coordinates allows multiple coordinates to be updated simultaneously, leading to a faster convergence rate.
To the best of our knowledge, Newton proposal makes the first attempt to utilize Newton's series expansion and multilinear extension in discrete sampling, which fills the gap of efficient sampling for complex discrete distributions when gradients are not available.
For different application scenarios, we develop several variants with Newton proposal, including unadjusted and Metropolis-adjusted versions. 
We theoretically prove the convergence and efficiency of Newton proposal without and with the MH step. 
Empirical results on various problems demonstrate the superiority of our method over baselines in general settings.

\section*{Acknowledgments}
We would like to thank Yingzhen Li for helpful discussions and the anonymous reviewers for their thoughtful comments on the manuscript.
\bibliography{references}

\newpage
\appendix
\onecolumn

\section{Detailed Derivation of Newton Proposal}
We give more details about the derivation of our Newton proposal.

We start from the MALA-like locally-balanced proposal~\citep{zanella2020informed} we mentioned in Section 3:
\begin{equation}\label{eq:Q_supp}
    q_0\left(\theta^{\prime} \mid \theta\right)=\frac{1}{Z_{\Theta}(\theta)}\exp\left(\frac{U(\theta^{\prime})-U(\theta)}{2}-\frac{\|\theta^{\prime}-\theta\|^2}{2\alpha}\right)
\end{equation}
where $Z_{\Theta}(\theta)$ is the normalizing constant.
When we use Newton's series expansion to approximate the local difference $U(\theta^{\prime})-U(\theta)$, we get
\begin{equation}\label{eq:square_supp}
\begin{aligned}
    q\left(\theta^{\prime} \mid \theta\right)=\widehat{q_0}\left(\theta^{\prime} \mid \theta\right)&=\frac{1}{Z_{\Theta}(\theta)}\exp\left(\frac{{\Delta_{h}[U](\theta)}^{\top}\cdot(\theta^{\prime}-\theta)}{2}-\frac{\|\theta^{\prime}-\theta\|^2}{2\alpha}\right)\\
    &\propto \exp{\left(-\frac{1}{2\alpha}\left((\theta^{\prime}-\theta)^2-\alpha\Delta_h[U](\theta)^{\top}\cdot(\theta^{\prime}-\theta)+\frac{\alpha^2}{4}\Delta_h[U](\theta)^2\right)\right)}\\
    &=\exp{\left(-\frac{1}{2\alpha}\|\theta^{\prime}-\theta-\frac{\alpha}{2}\Delta_h[U](\theta)\|^2\right)}
\end{aligned}
\end{equation}
where the second line is because $\Delta_h[U](\theta)$ is independent of $\theta^{\prime}$ and will not affect the normalized result.

Since \eqref{eq:square_supp} is actually an $\ell$-2 norm, $q\left(\theta^{\prime}\mid \theta\right)$ can be factorized coordinatewisely.
Besides, we assume the domain can be factorized coordinatewise.
Therefore, we can factorize $q\left(\theta^{\prime}\mid \theta\right)$ in \eqref{eq:square} as $q\left(\theta^{\prime} \mid \theta\right)=\prod_{i=1}^{d} q_{i}\left(\theta_{i}^{\prime} \mid \theta\right)$ and
\begin{equation}
\begin{aligned}
    q_{i}\left(\theta_{i}^{\prime} \mid \theta\right)&=\exp{\left(-\frac{1}{2\alpha}(\theta_i^{\prime}-\theta_i-\frac{\alpha}{2}\Delta_h[U](\theta))^2\right)}\\
    &\propto \exp{\left(\frac{1}{2}\Delta_h[U](\theta_i)(\theta_i^{\prime}-\theta_i)-\frac{(\theta_i^{\prime}-\theta_i)^2}{2\alpha}\right)}\\
    &=\operatorname{Softmax}\left(\frac{1}{2}\Delta_h[U](\theta_i)(\theta_i^{\prime}-\theta_i)-\frac{(\theta_i^{\prime}-\theta_i)^2}{2\alpha}\right)
    .
\end{aligned}
\end{equation}
Then we get our Newton proposal which is easy to compute in parallel:
\begin{equation}\label{eq:softmax}
    \operatorname{Categorical} \left(\operatorname{Softmax}\left(\frac{1}{2}\Delta_h[U](\theta_i)\left(\theta_{i}^{\prime}-\theta_{i}\right)-\frac{\left(\theta_{i}^{\prime}-\theta_{i}\right)^{2}}{2 \alpha}\right)\right).
\end{equation}

\section{Algorithm for Binary Variables}
When the variable domain $\Theta$ is binary $\left\{0, 1\right\}^d$, if we flip any coordinate $\theta_i$ to $\theta_i^{\prime}$, $(\theta_i^{\prime}-\theta_i)^2$ is always 1.
Thanks to the coordinatewise factorization, the sample space only contains 2 states: flipping or remaining the original state, which makes the normalizing constant $Z_{\Theta}(\theta)$ tractable.
In this way, we could simplify Algorithm 1 in the main body of our paper further and obtain the following algorithm, which clearly shows that our method can be cheaply computed in parallel on CPUs and GPUs.
We give the pseudo code when sampling from binary distributions with Newton proposal as follows.

\begin{algorithm}
\caption{Samplers with Newton Proposal on Binary Domains.}
\label{alg:binary}
\begin{algorithmic}[0]
\STATE \textbf{Input:} Stepsize $\alpha$.
\STATE \textbf{loop}
\STATE \qquad \textbf{Compute} 
$P(\theta)=\frac{\exp \left(-\frac{1}{2} \Delta_h[U](\theta) \odot(2 \theta-1)-\frac{1}{2 \alpha}\right)}{\exp \left(-\frac{1}{2} \Delta_h[U] (\theta) \odot(2 \theta-1)-\frac{1}{2 \alpha}\right)+1}$
\STATE \qquad \textbf{sample} $\mu \sim \text{Unif}(0,1)^d$
\STATE \qquad $I \leftarrow \operatorname{dim}(\mu \leq P(\theta))$
\STATE \qquad $\theta^{\prime} \leftarrow \operatorname {flipdim }(I)$

\STATE \qquad $\triangleright$ Optionally, do the MH step
\STATE \qquad \textbf{compute} $q\left(\theta^{\prime} \mid \theta\right)=\prod_{i} q_{i}\left(\theta_{i}^{\prime} \mid \theta\right)=\prod_{i \in I} P(\theta)_{i} \cdot \prod_{i \notin I}\left(1-P(\theta)_{i}\right)$
\STATE \qquad \textbf{compute} $P(\theta^{\prime})=\frac{\exp \left(-\frac{1}{2} \Delta_h[U](\theta^{\prime}) \odot(2 \theta^{\prime}-1)-\frac{1}{2 \alpha}\right)}{\exp \left(-\frac{1}{2} \Delta_h[U] (\theta^{\prime}) \odot(2 \theta^{\prime}-1)-\frac{1}{2 \alpha}\right)+1}$
\STATE \qquad \textbf{compute} $q\left(\theta \mid \theta^{\prime}\right)=\prod_{i} q_{i}\left(\theta_{i} \mid \theta^{\prime}\right)=\prod_{i \in I} P\left(\theta^{\prime}\right)_{i} \cdot \prod_{i \notin I}\left(1-P\left(\theta^{\prime}\right)_{i}\right)$
\STATE \qquad \textbf{set} $\theta \leftarrow \theta^{\prime}$ with probability 
$$
\min \left(1, \exp \left(U\left(\theta^{\prime}\right)-U(\theta)\right) \frac{q\left(\theta \mid \theta^{\prime}\right)}{q\left(\theta^{\prime} \mid \theta\right)}\right)
$$
\STATE \textbf{end loop}
\STATE \textbf{Output:} samples $\left\{\theta_k\right\}$.
\end{algorithmic}
\end{algorithm}

\section{Newton Proposal for Categorical Variables}
\subsection{Method 1: Newton's Series Approximation}
When using one-hot vectors to represent categorical variables, our Newton proposal becomes
\begin{equation}
    \text { Categorical }\left(\operatorname{Softmax}\left(\frac{1}{2} \Delta_h[U](\theta_i)^{\top}\left(\theta_i^{\prime}-\theta_i\right)-\frac{\left\|\theta_i^{\prime}-\theta_i\right\|_2^2}{2 \alpha}\right)\right),
\end{equation}
where $\theta_i, \theta_i^{\prime}$ are one-hot vectors.

If the variables are ordinal with clear ordering information, we can also use integer representation $\theta \in$ $\{0,1, \ldots, L-1\}^d$ and compute the Newton proposal as in Equation \eqref{eq:softmax}.
\subsection{Method 2: Multilinear Extension}
As we stated in Section 5.1, for binary variables, our Newton proposal obtained from the first-order Newton's series approximation is equivalent to that obtained from the first-order Taylor series approximation to the multilinear extension of the original discrete target distribution.
For categorical variables, we not only need to decide which coordinate to flip, but also need to decide the level.
By introducing the concepts of 'multiset', we generalize the multilinear extension to categorical distributions and thus extend our Newton proposal to categorical variables.

\subsubsection{Multiset}  
In classical sets, distinct elements can only occur once. 
A multiset is a natural generalization of a set, where elements can be contained repeatedly. 
The number of times an element occurs is called the multiplicity $\mu(i)$ of the element $i$.
A multiset $\mathcal{M}_{\mathcal{V}}$ is defined as a pair $\langle\mathcal{V}, \mu\rangle$, where $\mathcal{V}$ is the support and $\mu: \mathcal{V} \rightarrow \mathbb{N}$ is a function defining multiplicity for each element~\citep{sahin2020sets}. 
Given this definition, we can use the integer vector of the multiset's multiplicity to represent any multiset.
Also, we can transfer several important notions from multisets to integer vectors, such as the notion of a subset, set intersection, set union and set difference.

Now consider sampling from a $d$-dimensional discrete distribution.
The support can be represented as $\mathcal{X}=\prod_{i=1}^{d} \mathcal{X}_{i}$, where $\mathcal{X}_{i}=\{0,1, \ldots, L_i-1\}$ and the discrete function $f$ is an integer function defined as $f$: $\mathcal{X}\rightarrow \mathbb{N}$.
For ease of notation, we assume that $L_i$ does not depend on $i$ and $\mathcal{X}_{i}$ is the same along each dimension, $i.e.$, $\mathcal{X}_{i}=\{0,1, \ldots, L-1\}$, $\forall i\in {1,\ldots,d}$.
The obtained sample will be an integer vector $\bm{x}=(x_1,\ldots,x_n)$ ($x_i\in \mathcal{X}_{i}$, $\forall i\in {1,\ldots,n}$) and can be equivalently represented as a multiset.
Note that the integer $x_i$~($\forall i\in {1,\ldots,n}$) can be also represented as a binary vector $\bm{x}_i$ of length $L-1$. 
Take the $k=3$ case as an example.
$\bm{x}_i$ can be $(0,0)^{\top}$, $(1,0)^{\top}$ or $(0,1)^{\top}$, corresponding to the level of 0, 1, 2, respectively.
For simplicity in the calculation, we will use the binary representation in the following parts.

\subsubsection{Generalized Multilinear Extension}
Given the discrete distribution $f(\theta)$, the generalized multilinear extension will extend $f$ to a continuous domain while keeping the values on the original discrete domain.

Let $\bm{\rho}_{i} \in \mathbb{R}_{+}^{L-1}$ be the marginals of a $d$-dimensional categorical distribution and $\bm{\rho}:=\left[\bm{\rho}_{1} ; \ldots ; \bm{\rho}_{d}\right] \in \{0,1\}^{(L-1)\times d}$ is the concatenation of all $\bm{\rho}_{i}$ vectors. 
Each $\bm{\rho}_{i}$ lives in the $L-1$ dimensional simplex $\Delta^{L-1}$.
The simplex $\Delta^{L-1}$ is defined as
$$
\begin{array}{r}
\Delta^{L-1}:=\left\{\rho_{i} \in \mathbb{R}^{L-1}: \rho_{i, 1}+\ldots+\rho_{i, L-1} \leq 1\right. \\
\left.\rho_{i j} \geq 0, j=1, \ldots, L-1\right\}.
\end{array}
$$
We define the union of $n$ simplexes as $\Delta_{n}^{L-1}$, and naturally $\bm{\rho} \in \Delta_{d}^{L-1}$. 
Once we sample from $\bm{\rho}$, we get $\bm{\theta}=(\bm{\theta}_1,\ldots,\bm{\theta}_d)\in\{0,1\}^{(L-1)\times d}$.
The generalized multilinear extension $F$ is defined on the space of the product of categorical distributions and can be written as:

\begin{equation}\label{eq:integer}
    F\left(\bm{\rho}\right)=\mathbb{E}_{\bm{\theta} \sim \boldsymbol{\rho}_{1}, \ldots, \boldsymbol{\rho}_{d}}[f(\bm{\theta})].
\end{equation}
We need to compute the sum of $L^{d}$ elements to compute the expectation in Equation \eqref{eq:integer}. 
Note that when $L = 2$, this extension corresponds to the multilinear
extension of a set function.
Here is an example with $d=2$ and $L=3$. 
In this case, we have two categorical distributions which take three different values.

$$
\begin{aligned}
F(\bm{\rho})=&F\left(\left[\boldsymbol{\rho}_{1} ; \boldsymbol{\rho}_{2}\right]\right)=F\left(\rho_{11}, \rho_{12}, \rho_{21}, \rho_{22}\right)\\
=&f(\begin{pmatrix} 
0 & 0 \\ 0 & 0 
\end{pmatrix})\left(1-\rho_{11}-\rho_{12}\right)\left(1-\rho_{21}-\rho_{22}\right)
+f(\begin{pmatrix} 
0 & 0 \\ 1 & 1 
\end{pmatrix}) \rho_{12} \rho_{22}
+f(\begin{pmatrix} 
1 & 0 \\ 0 & 0
\end{pmatrix}) \rho_{11}\left(1-\rho_{21}-\rho_{22}\right)\\
+&f(\begin{pmatrix} 
0 & 1 \\ 0 & 0 
\end{pmatrix})\left(1-\rho_{11}-\rho_{12}\right) \rho_{21}
+f(\begin{pmatrix} 
0 & 0 \\ 1 & 0 
\end{pmatrix}) \rho_{12}\left(1-\rho_{21}-\rho_{22}\right)
+f(\begin{pmatrix} 
0 & 0 \\ 0 & 1 
\end{pmatrix})\left(1-\rho_{11}-\rho_{12}\right) \rho_{22}\\
+&f(\begin{pmatrix} 
1 & 1 \\ 0 & 0
\end{pmatrix}) \rho_{11} \rho_{21}+f(\begin{pmatrix} 
1 & 0 \\ 0 & 1
\end{pmatrix}) \rho_{11} \rho_{22}
+f(\begin{pmatrix} 
0 & 1 \\ 1 & 0 
\end{pmatrix}) \rho_{12} \rho_{21},
\end{aligned}
$$
where we have the following constraints
$$
\rho_{11}+\rho_{12} \leq 1, \rho_{21}+\rho_{22} \leq 1, \rho_{i j} \geq 0, i, j=1,2 .
$$

\subsubsection{Newton Proposal via Generalized Multilinear Extension}
Similar to the multilinear extension for the binary domain, we can calculate the first-order partial derivative of the generalized multilinear extension of $f$, $i.e.$, $F\left(\bm{\rho}_{1}, \ldots, \boldsymbol{\rho}_{d}\right)$.
Given $\bm{\rho} \in \Delta_{d}^{L-1}$, let $\mathcal{R}_{\mathcal{V}}$ be a random multiset where elements appear independently with probabilities $\boldsymbol{\rho}_{i}$. Since $F$ is multilinear, the partial derivative can be written as a difference of two generalized multilinear extensions: 
$$
\begin{aligned}
\frac{\partial F}{\partial \rho_{i j}} &=F\left(\boldsymbol{\rho}_{1}, \boldsymbol{\rho}_{i}=\bm{e}_{j}, \boldsymbol{\rho}_{n}\right)-F\left(\boldsymbol{\rho}_{1}, \boldsymbol{\rho}_{i}=\mathbf{0}, \boldsymbol{\rho}_{n}\right) \\
&=\mathbb{E}_{\mathcal{R}_{\mathcal{V} \sim \rho}}\left[f\left(\mathcal{R}_{\mathcal{V}}\cup\mathcal{E}_{i}^{j}\right)\right]-\mathbb{E}_{\mathcal{R}_{\mathcal{V} \sim \boldsymbol{\rho}}}\left[f\left(\mathcal{R}_{\mathcal{V}}\setminus\mathcal{E}_{i}^{j}\right)\right] \\
\end{aligned}
$$

where $\cup$ and $\setminus$ corresponds to union and set difference between multisets, respectively.
$\bm{e}_j\in \Delta_{n}^{k-1}$ is a unit vector with the $j$-th element 1.
$\mathcal{E}_{i}^{j}$ is the multiset whose $i$-th element has multiplicity $j$.
In this way, we can get $\nabla{F}\in\mathbb{R}^{(L-1)\times d}$ and calculate the proposal of each coordinate as below:
\begin{equation}\label{eq:categorical}
\text{Categorical} \left(\operatorname{Softmax}\left(\frac{1}{2} \nabla F({\theta})_{i}^{\top}\left({\theta}_{i}^{\prime}-{\theta}_{i}\right)-\frac{\|{\theta}_{i}^{\prime}-{\theta}_{i}\|_2^{2}}{2 \alpha}\right)\right).
\end{equation}

\subsection{Experiments}
We implement the Newton proposal on the dim$=4\times 4\times 3\times3$ Potts model.
\begin{figure}[htbp]
\vspace{-0.3cm}
\centering
\includegraphics[height=5cm]{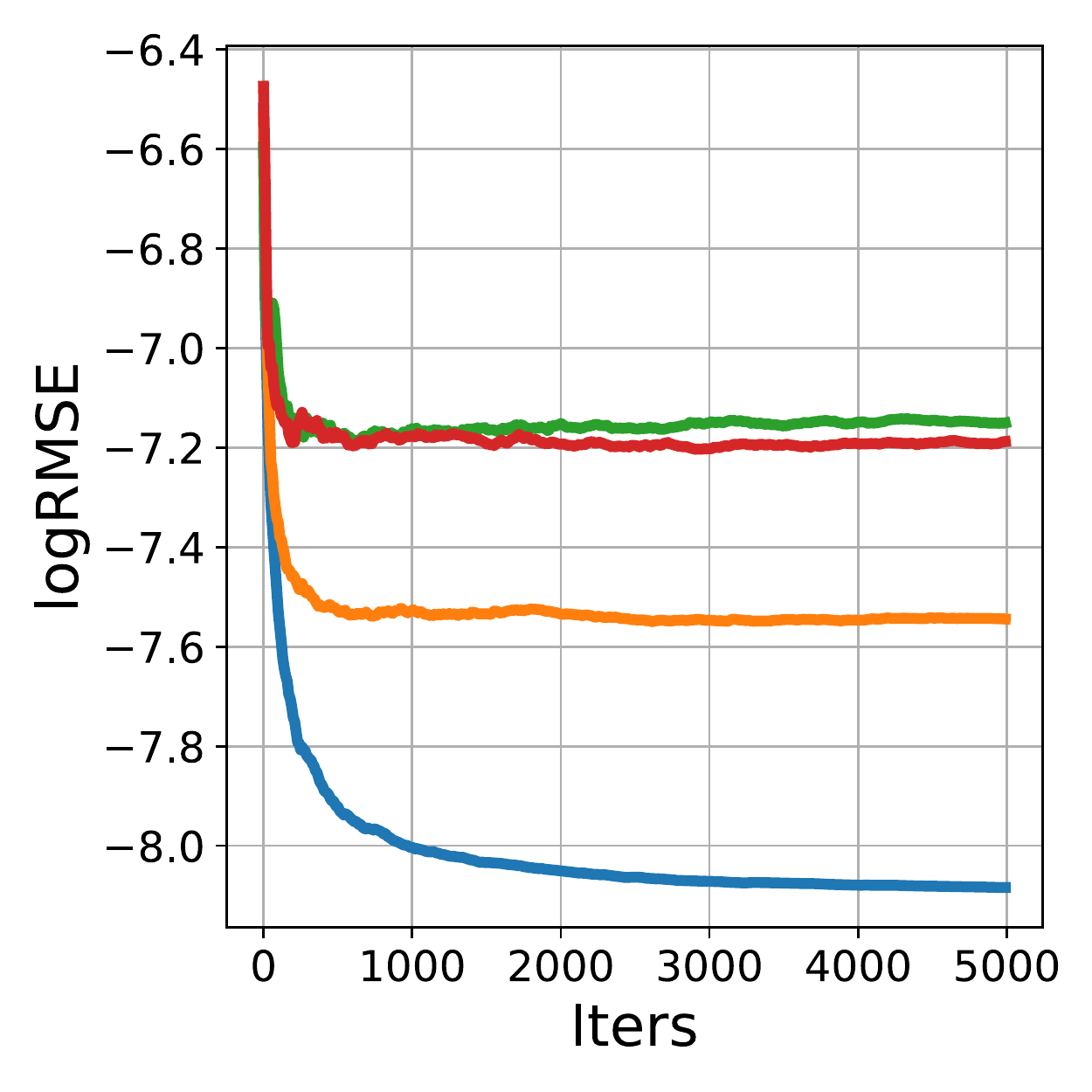}\label{fig:potts_iter}
\includegraphics[height=5cm]{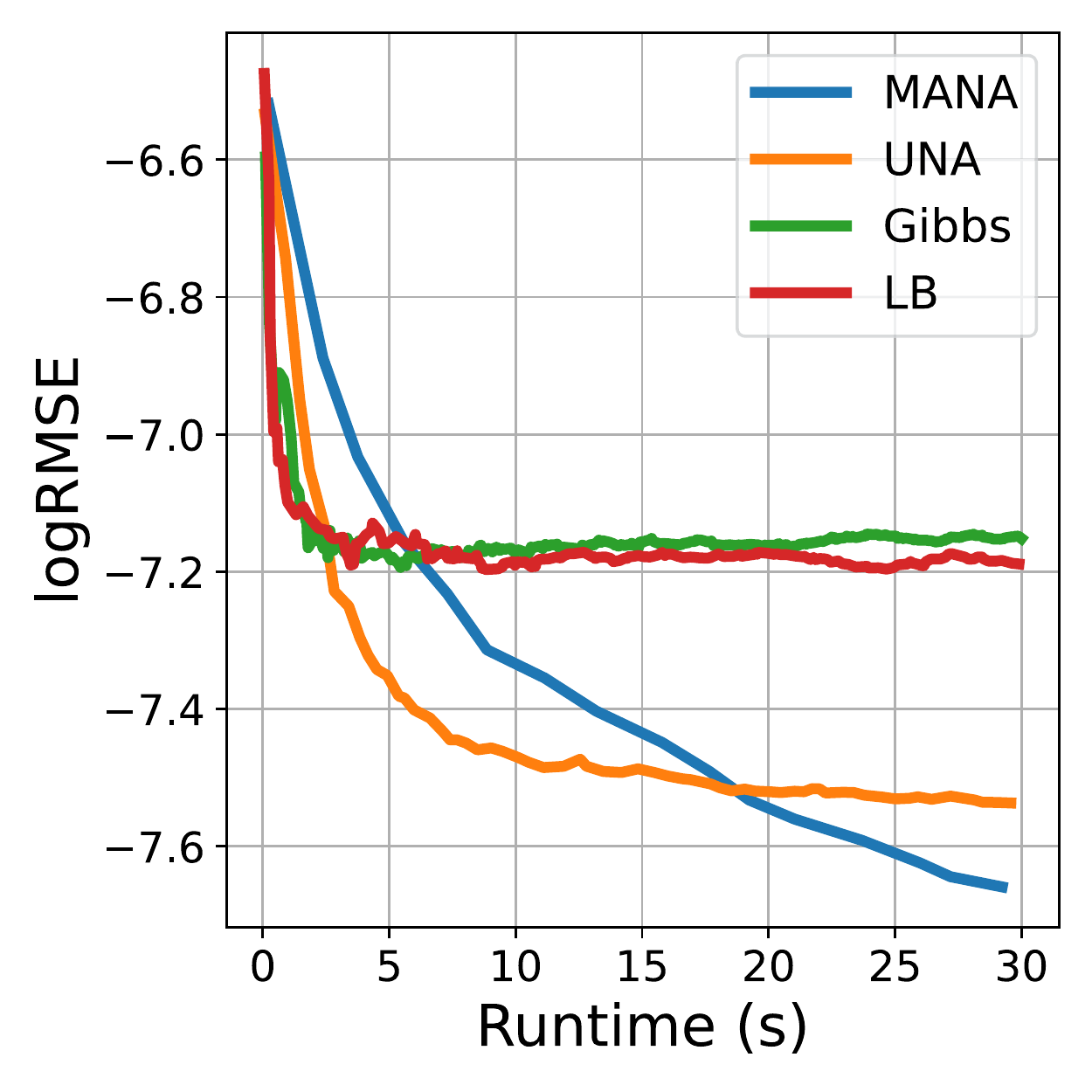}\label{fig:potts_time}
\caption{Potts model: lines of MANA run under others.}
\label{fig:potts}
\vspace{-0.4cm}
\end{figure}

Figure \ref{fig:potts} shows that MANA converges fast in both the number of iterations and running time.
Our Newton proposal achieves promising performance for categorical variables.

\section{Details and Proof of Theorem 1}
As we discuss in Section 5, our Newton proposal can be equivalently obtained by conducting Taylor expansion on the multilinear extension of the target discrete distribution.
In this section, we focus on second-order modular functions and study the asymptotic convergence of UNA.
\subsection{Definitions of Submodular Function and Multilinear Extension}
A \textbf{submodular function} is a set function whose value has the property that the difference in the incremental value of the function that a single element makes when added to an input set decreases as the size of the input set increases. 
In a word, submodular functions have a natural diminishing returns property.
If $f$ is submodular then its multilinear extension, $i.e.$, $F$, is concave along any line $d \geq 0$.

For a discrete distribution whose function is a set function $f: 2^{D} \rightarrow \mathbb{R}$, its \textbf{multilinear extension} $F:[0,1]^{d} \rightarrow \mathbb{R}$ is defined as:
\begin{equation}\label{eq:multilinear_supp}
    F({\theta})=\sum_{S \subseteq D} f(S) \prod_{i \in S} \theta_{i} \prod_{i \in D \backslash S}\left(1-\theta_{i}\right).
\end{equation}
It is possible to relate properties of $f$ to properties of its multilinear extension $F$.
In particular, we have:

\textbf{Proposition 1.} Let $F$ be the multilinear extension of $f$, then:
\begin{enumerate}
    \item If $f$ is non-decreasing, then $F$ is non-decreasing along any direction $d \geq 0$.
    \item If $f$ is submodular then $F$ is concave along any line $d \geq 0$.
\end{enumerate}
Both properties can be established by first looking at how $F$ behaves along coordinates axes.
We first calculate the first and second order derivative of $F({\theta})$.
\begin{enumerate}
    \item Let $i \in D$, since $F$ is linear in $\theta_{i}$, we have:\\
\begin{eqnarray}\label{eq:1der_supp}
\frac{\partial F}{\partial \theta_{i}}({\theta})=&F\left(\theta_{1}, \ldots, \theta_{i-1}, 1, \theta_{i+1}, \ldots, \theta_{d}\right)-F\left(\theta_{1}, \ldots, \theta_{i-1}, 0, \theta_{i+1}, \ldots, \theta_{d}\right)\nonumber
\end{eqnarray}
Let $R$ be the random subset of $D \backslash\{i\}$ where each element $j \in D \backslash\{i\}$ is included with probability $\theta_{j}$, then we can rewrite:
\begin{eqnarray}\label{eq:order1}
\frac{\partial F}{\partial \theta_{i}}({\theta})=\mathbb{E}[f(R \cup\{i\})]-\mathbb{E}[f(R)].
\end{eqnarray}

\item Similarly, let $R_2$ be the random subset of $D \backslash\{i, j\}$ and $R_3$ be the random subset of $D \backslash\{j\}$ where each element $k$ is included with probability $\theta_{k}$, we have:
\begin{eqnarray}
\begin{aligned}\label{eq:sec-derivative}
\frac{\partial^{2} F}{\partial \theta_{i} \partial \theta_{j}}({\theta})&=\mathbb{E}[f(R_2 \cup\{i, j\})]-\mathbb{E}[f(R_2 \cup\{i\})]-\mathbb{E}[f(R_2 \cup\{j\})]+\mathbb{E}[f(R_2)]\\
&=\mathbb{E}[\big(f(R_2 \cup\{i, j\})-f(R_2 \cup\{i\})\big)-\big(f(R_2 \cup\{j\})-f(R_2)\big)].
\end{aligned}
\end{eqnarray}
By submodularity of $f$, the last quantity in \eqref{eq:sec-derivative} is non-positive, $i.e.$, $\frac{\partial^2 F}{\partial \theta_i \partial \theta_j}(\theta) \leq 0$.
\end{enumerate}

Let $\theta \in[0,1]^n$ and $d \geq 0$. We define the function $F_{\theta, d}(\lambda)=F(\theta+\lambda d)$ of the real variable $\lambda$. We note that $F_{\theta, d}^{\prime}(\lambda)=\langle d, \nabla F(\theta+\lambda d)\rangle$ and $F_{\theta, d}^{\prime \prime}=d^T H_f(\theta+\lambda d) d$.
\begin{enumerate}
    \item If $f$ is non-decreasing, then $\nabla F(\theta+\lambda d) \geq 0$ and $\langle d, \nabla F(\theta+\lambda d)\rangle \geq 0$. Hence $F_{\theta, d}$ is nondecreasing.
    \item If $f$ is submodular, then $H_f(\theta+\lambda d) \leq 0$ and $d^T H_f(\theta+\lambda d) d \leq 0$. Hence $F_{\theta, d}$ is concave.
\end{enumerate}

\subsection{Second-Order Modular}
For a submodular function $f$, let $M G(A, e)=f(A \cup\{e\})-$ $f(A)$ denote the marginal gain from adding element $e$ to set $A$. For sets $A, S$, we define $G R(A, S, e)=M G(A, e)-M G(A \cup S, e)$ as the amount by which $S$ reduces the marginal gain from adding e to $A$. (Here, GR stands for Gain Reduction.) Note that by definition of submodularity, $G R(A, S, e)$ is always non-negative.

The function $f$ is said to be \textbf{second-order modular} if, for all sets $A, B, S$ such that $A \subseteq B$, and $S \cap B=\emptyset$, and all elements $e$, we have: $G R(A, S, e)=$ $G R(B, S, e)$~\citep{korula2018online}.

Specifically, if for any set $R_2$, $\forall i, j$, $f(R_2 \cup\{i,j\})-f(R_2\cup\{i\})=b_j$ and $f(R_2 \cup\{j\})-f(R_2)=A_{ij}+b_j$ where $A_{ij}$ and $b_j$ are both constants, then $GR(R_2,\{i\},j)=b_j$.
At this time, $f$ is second-order modular.
That is to say, the second-order modular function is in the form of $f(D) = \sum_{u,v\in D}A(u,v)+\sum_{u\in D} b(u)$, whose multilinear function is 
\begin{eqnarray}\label{eq:multi}
F(\theta)=\theta^{\top} A \theta+b \theta,
\end{eqnarray}
which is exactly in the same form with the energy function of Ising model.

\subsection{Proof of Theorem 1}
\begin{proof}
We finish the proof in the view of multiliear extension, $i.e.$, we see the multilinear extension of the original discrete distribution as the energy function.
We first prove the weak convergence and then prove the convergence rate with respect to the stepsize $\alpha$.

\textbf{(1) Weak convergence.} 
When $f(D)$ is second-order modular, the Hessian matrix of its multilinear extension $F$ in Equation \eqref{eq:multi} will be a constant, $i.e.$, $\frac{\partial^{2} F}{\partial \theta_{i} \partial \theta_{j}}({\theta}) = A_{ij},~\forall {\theta}$.
We have that $\nabla F(\theta)=2 A^{\top} \theta+b, \nabla^{2} F(\theta)=2 A$. Since $\nabla^{2} F(\theta)$ is a constant, we can rewrite the proposal distribution as the following
$$
\begin{aligned}
q_\alpha\left(\theta^{\prime} \mid \theta\right) &=\frac{\exp \left(\frac{1}{2} \nabla F(\theta)^{\top}\left(\theta^{\prime}-\theta\right)-\frac{1}{2 \alpha}\left\|\theta^{\prime}-\theta\right\|^2\right)}{\sum_x \exp \left(\frac{1}{2} \nabla F(\theta)^{\top}(x-\theta)-\frac{1}{2 \alpha}\|x-\theta\|^2\right)} \\
&=\frac{\exp \left(\frac{1}{2} \nabla F(\theta)^{\top}\left(\theta^{\prime}-\theta\right)+\frac{1}{2}\left(\theta^{\prime}-\theta\right)^{\top} A\left(\theta^{\prime}-\theta\right)-\left(\theta^{\prime}-\theta\right)^{\top}\left(\frac{1}{2 \alpha} I+\frac{1}{2} A\right)\left(\theta^{\prime}-\theta\right)\right)}{\sum_x \exp \left(\frac{1}{2} \nabla F(\theta)^{\top}(x-\theta)+\frac{1}{2}(x-\theta)^{\top} A(x-\theta)-(x-\theta)^{\top}\left(\frac{1}{2 \alpha} I+\frac{1}{2} A\right)(x-\theta)\right)} \\
&=\frac{\exp \left(\frac{1}{2}\left(F\left(\theta^{\prime}\right)-F(\theta)\right)-\left(\theta^{\prime}-\theta\right)^{\top}\left(\frac{1}{2 \alpha} I+\frac{1}{2} A\right)\left(\theta^{\prime}-\theta\right)\right)}{\sum_x \exp \left(\frac{1}{2}(F(x)-F(\theta))-(x-\theta)^{\top}\left(\frac{1}{2 \alpha} I+\frac{1}{2} A\right)(x-\theta)\right)}
\end{aligned}
$$
where the last equation is because the Taylor expansion
$F\left(\theta^{\prime}\right)-F(\theta)=\nabla F(\theta)^{\top}\left(\theta^{\prime}-\theta\right)+\frac{1}{2}\left(\theta^{\prime}-\theta\right)^{\top} 2 A\left(\theta^{\prime}-\theta\right)$.

Let $Z_\alpha(\theta)=\sum_x \exp \left(\frac{1}{2}(F(x)-F(\theta))-(x-\theta)^{\top}\left(\frac{1}{2 \alpha} I+\frac{1}{2} A\right)(x-\theta)\right)$, and $\pi_\alpha=\frac{Z_\alpha(\theta) \pi(\theta)}{\sum_x Z_\alpha(x) \pi(x)}$, now we will show that $q_\alpha$ is reversible w.r.t. $\pi_\alpha$.
We have that
\begin{eqnarray}\label{eq:pq}
\begin{aligned}
\pi_\alpha(\theta) q_\alpha\left(\theta^{\prime} \mid \theta\right) &=\frac{Z_\alpha(\theta) \pi(\theta)}{\sum_x Z_\alpha(x) \pi(x)} \cdot \frac{\exp \left(\frac{1}{2}\left(F\left(\theta^{\prime}\right)-F(\theta)\right)-\left(\theta^{\prime}-\theta\right)^{\top}\left(\frac{1}{2 \alpha} I+\frac{1}{2} A\right)\left(\theta^{\prime}-\theta\right)\right)}{Z_\alpha(\theta)} \\
&=\frac{\exp \left(\frac{1}{2}\left(F\left(\theta^{\prime}\right)+F(\theta)\right)-\left(\theta^{\prime}-\theta\right)^{\top}\left(\frac{1}{2 \alpha} I+\frac{1}{2} A\right)\left(\theta^{\prime}-\theta\right)\right)}{Z \cdot \sum_x Z_\alpha(x) \pi(x)} .
\end{aligned}
\end{eqnarray}
We can see that the expression in \eqref{eq:pq} is symmetric in $\theta$ and $\theta^{\prime}$. Therefore $q_\alpha$ is reversible and the stationary distribution is $\pi_\alpha$.
Now we will prove that $\pi_\alpha$ converges weakly to $\pi$ as $\alpha \rightarrow 0$. Notice that for any $\theta$,
$$
\begin{aligned}
Z_\alpha(\theta) &=\sum_x \exp \left(\frac{1}{2}(F(x)-F(\theta))-(x-\theta)^{\top}\left(\frac{1}{2 \alpha} I+\frac{1}{2} A\right)(x-\theta)\right) \\
& \stackrel{\alpha \downarrow 0}{\Longrightarrow} \sum_x \exp \left(\frac{1}{2}(F(x)-F(\theta))\right) \delta_\theta(x) \\
&=1,
\end{aligned}
$$
where $\delta_\theta(x)$ is a Dirac delta. It follows that $\pi_\alpha$ converges pointwisely to $\pi(\theta)$. By Scheffé's Lemma, we attain that $\pi_\alpha$ converges weakly to $\pi$.

\textbf{(2) Convergence Rate w.r.t. Stepsize.} 

Let us consider the convergence rate in terms of $L_{1}$-norm
$$
\left\|\pi_\alpha-\pi\right\|_1=\sum_\theta\left|\frac{Z_\alpha(\theta) \pi(\theta)}{\sum_x Z_\alpha(x) \pi(x)}-\pi(\theta)\right|.
$$
We write out each absolute value term
$$
\begin{aligned}
\left|\frac{Z_\alpha(\theta) \pi(\theta)}{\sum_x Z_\alpha(x) \pi(x)}-\pi(\theta)\right| &=\pi(\theta)\left|\frac{Z_\alpha(\theta)}{\sum_x Z_\alpha(x) \pi(x)}-1\right| \\
&=\pi(\theta)\cdot\\
&~\left|\frac{1+\sum_{x \neq \theta} \exp \left(\frac{1}{2} F(x)-\frac{1}{2} F(\theta)-(x-\theta)^{\top}\left(\frac{1}{2 \alpha} I+\frac{1}{2} A\right)(x-\theta)\right)}{1+\sum_y \frac{1}{Z} \exp (F(y)) \sum_{x \neq y} \exp \left(\frac{1}{2} F(x)-\frac{1}{2} F(y)-(x-y)^{\top}\left(\frac{1}{2 \alpha} I+\frac{1}{2} A\right)(x-y)\right)}-1\right| .
\end{aligned}
$$

Since $\lambda_{\min }(A)\|x\|^2 \leq x^{\top} A x, \forall x$, it follows that
$$
(x-\theta)^{\top}\left(\frac{1}{2 \alpha} I+\frac{1}{2} A\right)(x-\theta) \geq \frac{1+\alpha \lambda_{\min }}{2 \alpha}\|x-\theta\|^2 .
$$
We also notice that $\min _{x \neq \theta}\|x-\theta\|^2=1$, thus when $\frac{Z_\alpha(\theta)}{\sum_x Z_\alpha(x) \pi(x)}-1>0$, we get
$$
\begin{aligned}
\left|\frac{Z_\alpha(\theta) \pi(\theta)}{\sum_x Z_\alpha(x) \pi(x)}-\pi(\theta)\right| &=\pi(\theta)\cdot\\
&~\left(\frac{1+\sum_{x \neq \theta} \exp \left(\frac{1}{2} F(x)-\frac{1}{2} F(\theta)-(x-\theta)^{\top}\left(\frac{1}{2 \alpha} I+\frac{1}{2} A\right)(x-\theta)\right)}{1+\sum_y \frac{1}{Z} \exp (F(y)) \sum_{x \neq y} \exp \left(\frac{1}{2} F(x)-\frac{1}{2} F(y)-(x-y)^{\top}\left(\frac{1}{2 \alpha} I+\frac{1}{2} A\right)(x-y)\right)}-1\right) \\
& \leq \pi(\theta)\left(1+\sum_{x \neq \theta} \exp \left(\frac{1}{2} F(x)-\frac{1}{2} F(\theta)-\frac{1+\alpha \lambda_{\min }}{2 \alpha}\|x-\theta\|^2\right)-1\right) \\
& \leq \pi(\theta)\left(1+\exp \left(-\frac{1+\alpha \lambda_{\min }}{2 \alpha}\right) \sum_{x \neq \theta} \exp \left(\frac{1}{2} F(x)-\frac{1}{2} F(\theta)\right)-1\right) \\
&=\pi(\theta)\left(\sum_{x \neq \theta} \exp \left(\frac{1}{2} F(x)-\frac{1}{2} F(\theta)\right)\right) \cdot \exp \left(-\frac{1+\alpha \lambda_{\min }}{2 \alpha}\right) \\
& \leq \pi(\theta)\left(\sum_x \exp (F(x))\right) \cdot \exp \left(-\frac{1+\alpha \lambda_{\min }}{2 \alpha}\right) \\
&=\pi(\theta) Z \cdot \exp \left(-\frac{1+\alpha \lambda_{\min }}{2 \alpha}\right) .
\end{aligned}
$$

Similarly, when $\frac{Z_\alpha(\theta)}{\sum_x Z_\alpha(x) \pi(x)}-1<0$, we have,
$$
\begin{aligned}
\left|\frac{Z_\alpha(\theta) \pi(\theta)}{\sum_x Z_\alpha(x) \pi(x)}-\pi(\theta)\right| &=\pi(\theta)\cdot\\
&~\left(1-\frac{1+\sum_{x \neq \theta} \exp \left(\frac{1}{2} F(x)-\frac{1}{2} F(\theta)-(x-\theta)^{\top}\left(\frac{1}{2 \alpha} I+\frac{1}{2} A\right)(x-\theta)\right)}{1+\sum_y \frac{1}{Z} \exp (F(y)) \sum_{x \neq y} \exp \left(\frac{1}{2} F(x)-\frac{1}{2} F(y)-(x-y)^{\top}\left(\frac{1}{2 \alpha} I+\frac{1}{2} A\right)(x-y)\right)}\right) \\
& \leq \pi(\theta)\left(1-\frac{1}{1+\sum_y \frac{1}{Z} \exp (F(y)) \sum_{x \neq y} \exp \left(\frac{1}{2} F(x)-\frac{1}{2} F(y)-\frac{1+\alpha \lambda_{\min }}{2 \alpha}\right)}\right) \\
&=\pi(\theta)\left(\frac{\sum_y \frac{1}{Z} \exp (F(y)) \sum_{x \neq y} \exp \left(\frac{1}{2} F(x)-\frac{1}{2} F(y)-\frac{1+\alpha \lambda_{\min }}{2 \alpha}\right)}{1+\sum_y \frac{1}{Z} \exp (F(y)) \sum_{x \neq y} \exp \left(\frac{1}{2} F(x)-\frac{1}{2} F(y)-\frac{1+\alpha \lambda_{\min }}{2 \alpha}\right)}\right) \\
& \leq \pi(\theta)\left(\sum_y \frac{1}{Z} \exp (F(y)) \sum_{x \neq y} \exp \left(\frac{1}{2} F(x)-\frac{1}{2} F(y)\right)\right) \cdot \exp \left(-\frac{1+\alpha \lambda_{\min }}{2 \alpha}\right) \\
& \leq \pi(\theta)\left(\sum_x \exp (F(x))\right) \cdot \exp \left(-\frac{1+\alpha \lambda_{\min }}{2 \alpha}\right) \\
&=\pi(\theta) Z \cdot \exp \left(-\frac{1+\alpha \lambda_{\min }}{2 \alpha}\right) .
\end{aligned}
$$
Therefore, the difference between $\pi_\alpha$ and $\pi$ can be bounded as follows
$$
\left\|\pi_\alpha-\pi\right\|_1 \leq \sum_\theta \pi(\theta) Z \cdot \exp \left(-\frac{1+\alpha \lambda_{\min }}{2 \alpha}\right)=Z \cdot \exp \left(-\frac{1+\alpha \lambda_{\min }}{2 \alpha}\right) .
$$
\end{proof}

\section{Proof of Theorem 2}
\begin{proof}
Our proof follows from Theorem 1 of \citep{grathwohl2021oops} and Theorem 2 of \citep{zanella2020informed}, which state that for two $p$-reversible Markov transition kernels $Q_1\left(x^{\prime}, x\right)$ and $Q_2\left(x^{\prime}, x\right)$, if there exists $c>0$ for all $x^{\prime} \neq x$ such that $Q_1\left(x^{\prime}, x\right)>c \cdot Q_2\left(x^{\prime}, x\right)$ then
\begin{enumerate}
    \item $\operatorname{var}_p\left(h, Q_1\right) \leq \frac{\operatorname{var}_p\left(h, Q_1\right)}{c}+\frac{1-c}{c} \cdot \operatorname{var}_p(h)$
    \item $\operatorname{Gap}\left(Q_1\right) \geq c \cdot \operatorname{Gap}\left(Q_2\right)$
    \end{enumerate}
where $\operatorname{var}_p(h, Q)$ is the asymptotic variance and $\operatorname{Gap}(Q)$ is the spectral gap, which are both defined in the main body of this paper.
$\operatorname{var}_p(h)$ is the standard variance $E_p\left[h(x)^2\right]-E_p[h(x)]^2$.
Our proof proceeds by showing we can bound $Q^{\nabla}\left(x^{\prime}, x\right) \geq c \cdot Q\left(x^{\prime}, x\right)$, and the results of the theorem then follow directly from Theorem 2 of \citep{zanella2020informed}.

\subsection{Definitions}
We begin by writing down the proposal distribution of interest and their corresponding Markov transition kernels. 
For ease of notion we define some values
$$
\begin{aligned}
\Delta\left(\theta^{\prime}, \theta\right) &:=U\left(\theta^{\prime}\right)-U(\theta); \\
\tilde{\Delta}\left(\theta^{\prime}, \theta\right) &:=\Delta_h[U](\theta)^{\top}\left(\theta^{\prime}-\theta\right); \\
D &:=\sup _{\theta^{\prime} \in \Theta}\left\|\theta^{\prime}-\theta\right\|.
\end{aligned}
$$
Then our original proposal, $i.e.$, the MALA-like locally-balanced proposal for $\theta^{\prime}$ is
\begin{eqnarray}\label{eq:delta_proposal}
q_0\left(\theta^{\prime} \mid \theta\right)=\frac{\exp \left(\frac{1}{2}\Delta\left(\theta^{\prime}, \theta\right)-\frac{1}{2\alpha}(\theta^{\prime}-\theta)^2\right)}{Z(\theta)}
\end{eqnarray}

where we have defined
$$
Z(\theta)=\sum_{\theta^{\prime} \in \Theta} \exp \left(\frac{1}{2}\Delta\left(\theta^{\prime}, \theta\right)-\frac{1}{2\alpha}(\theta^{\prime}-\theta)^2\right).
$$

When we examine the acceptance rate of the proposal we find
$$
\begin{aligned}
&\exp \left(U\left(\theta^{\prime}\right)-U(\theta)\right) \frac{q_0\left(\theta \mid \theta^{\prime}\right)}{q_0\left(\theta^{\prime} \mid \theta\right)} \\
=&\exp \left(\Delta(\theta^{\prime}, \theta)\right) \frac{\exp \left(\frac{1}{2}\Delta(\theta,\theta^{\prime}\right)-\frac{1}{2\alpha}(\theta-\theta^{\prime})^2 )Z(\theta)}{\exp \left(\frac{1}{2}\Delta(\theta^{\prime},\theta)-\frac{1}{2\alpha}(\theta^{\prime}-\theta)^2\right) Z(\theta^{\prime})} \\
=&\exp(\frac{1}{2}\Delta(\theta^{\prime}, \theta)+\frac{1}{2}\Delta(\theta, \theta^{\prime}))\frac{Z(\theta)}{Z(\theta^{\prime})}\\
=&\frac{Z(\theta)}{Z\left(\theta^{\prime}\right)}
\end{aligned}
$$

Then the acceptance rate of the target proposal in \eqref{eq:delta_proposal} can be simplified as
$$
\begin{aligned}
\min \left\{1, \exp \left(U\left(\theta^{\prime}\right)-U(\theta)\right) \frac{q_0\left(\theta \mid \theta^{\prime}\right)}{q_0\left(\theta^{\prime} \mid \theta\right)}\right\}=\min\left\{1,\frac{Z(\theta)}{Z\left(\theta^{\prime}\right)}\right\}.
\end{aligned}
$$

This corresponding Markov transition kernel is
$$
\begin{aligned}
Q\left(\theta^{\prime}, \theta\right)&=q_0\left(\theta^{\prime} \mid \theta\right) \min \left\{1, \frac{Z(\theta)}{Z\left(\theta^{\prime}\right)}\right\} \\
&=\min \left\{\frac{\exp \left(\frac{1}{2}\Delta\left(\theta^{\prime}, \theta\right)-\frac{1}{2\alpha}(\theta^{\prime}-\theta)^2\right)}{Z(\theta)}, \frac{\exp \left(\frac{1}{2}\Delta\left(\theta^{\prime}, \theta\right)-\frac{1}{2\alpha}(\theta^{\prime}-\theta)^2\right)}{Z\left(\theta^{\prime}\right)}\right\}.
\end{aligned}
$$

Our proposed Newton proposal is the first-order Newton's series approximation of the original target proposal~\eqref{eq:delta_proposal} for $\theta^{\prime} \in \Theta$:
$$
q\left(\theta^{\prime} \mid \theta\right)=\frac{\exp \left(\frac{1}{2}\tilde{\Delta}\left(\theta^{\prime}, \theta\right)-\frac{1}{2\alpha}(\theta^{\prime}-\theta)^2\right)}{\tilde{Z}(\theta)}
$$
where we have defined
$$
\tilde{Z}(\theta)=\sum_{\theta^{\prime} \in \Theta} \exp \left(\frac{1}{2}\tilde{\Delta}\left(\theta^{\prime}, \theta\right)-\frac{1}{2\alpha}(\theta^{\prime}-\theta)^2\right).
$$

For our Newton proposal, we simplify the term in the acceptance rate of the proposal as
$$
\begin{aligned}
\exp \left(U\left(\theta^{\prime}\right)-U(\theta)\right) \frac{q\left(\theta \mid \theta^{\prime}\right)}{q\left(\theta^{\prime} \mid \theta\right)}=&\exp{(\Delta(\theta^{\prime},\theta))}\frac{\exp \left(\frac{1}{2}\tilde{\Delta}(\theta,\theta^{\prime})-\frac{1}{2\alpha}(\theta-\theta^{\prime})^2 \right)\tilde{Z}(\theta)}{\exp \left(\frac{1}{2}\tilde{\Delta}(\theta^{\prime},\theta)-\frac{1}{2\alpha}(\theta^{\prime}-\theta)^2\right) \tilde{Z}(\theta^{\prime})}\\
=&\exp\left(\Delta(\theta^{\prime},\theta)+\frac{1}{2}\tilde{\Delta}(\theta,\theta^{\prime})-\frac{1}{2}\tilde{\Delta}(\theta^{\prime},\theta)\right) \frac{\tilde{Z}(\theta)}{\tilde{Z}\left(\theta^{\prime}\right)}
\end{aligned}
$$
Then the Markov transition kernel
$$
\begin{aligned}
\tilde{Q}\left(\theta^{\prime}, \theta\right)
&=q\left(\theta^{\prime} \mid \theta\right) \min \left\{1, \exp\left(\Delta(\theta^{\prime},\theta)+\frac{1}{2}\tilde{\Delta}(\theta,\theta^{\prime})-\frac{1}{2}\tilde{\Delta}(\theta^{\prime},\theta)\right) \frac{\tilde{Z}(\theta)}{\tilde{Z}\left(\theta^{\prime}\right)}\right\} \\
&=\min \left\{\frac{\exp \left(\frac{1}{2}\tilde{\Delta}\left(\theta^{\prime}, \theta\right)-\frac{1}{2\alpha}(\theta^{\prime}-\theta)^2\right)}{\tilde{Z}(\theta)},
\frac{\exp\left(\Delta(\theta^{\prime},\theta)+\frac{1}{2}\tilde{\Delta}(\theta,\theta^{\prime})-\frac{1}{2\alpha}(\theta^{\prime}-\theta)^2\right)}{\tilde{Z}(\theta^{\prime})}\right\}.
\end{aligned}
$$

\subsection{Preliminaries}
It can be seen that $\tilde{\Delta}\left(\theta^{\prime}, \theta\right)$ is a first-order Newton's series approximation to $\Delta_h\left(\theta^{\prime}, \theta\right)$.
When the finite difference $\Delta[U](\theta)$ has an analog of Lipschitz continuity, $i.e.$, $\left|\tilde{\Delta}\left(\theta^{\prime}, \theta\right)-\Delta\left(\theta^{\prime}, \theta\right)\right| \leq \frac{L}{2}\left\|\theta^{\prime}-\theta\right\|^2$, since $\left\|\theta^{\prime}-\theta\right\|^2$ is bounded, we have
$$
-\frac{L}{2} D^2 \leq \tilde{\Delta}\left(\theta^{\prime}, \theta\right)-\Delta\left(\theta^{\prime}, \theta\right) \leq \frac{L}{2} D^2
$$

\subsection{Normalizing Constant Bounds}
We derive upper- and lower-bounds on $\tilde{Z}(\theta)$ in terms of $Z(\theta)$.
$$
\begin{aligned}
\tilde{Z}(\theta) &= \sum_{\theta^{\prime} \in \Theta} \exp \left(\frac{1}{2}\tilde{\Delta}\left(\theta^{\prime}, \theta\right)-\frac{1}{2\alpha}(\theta^{\prime}-\theta)^2\right)\\
& = \sum_{\theta^{\prime} \in \Theta}\exp\left(\frac{1}{2}{\Delta}\left(\theta^{\prime}, \theta\right)-\frac{1}{2\alpha}(\theta^{\prime}-\theta)^2\right)\cdot\exp\left(\frac{1}{2}\tilde{\Delta}(\theta^{\prime}, \theta)-\frac{1}{2}\Delta(\theta^{\prime}, \theta) \right)\\
& \leq \sum_{\theta^{\prime} \in \Theta} \exp \left(\frac{1}{2}\Delta\left(\theta^{\prime}, \theta\right)-\frac{1}{2\alpha}(\theta^{\prime}-\theta)^2\right) \cdot\exp \left(\frac{L D^2}{4}\right) \\
&=\exp \left(\frac{L D^2}{4}\right)\sum_{\theta^{\prime} \in \Theta} \exp \left(\frac{1}{2}\Delta\left(\theta^{\prime}, \theta\right)-\frac{1}{2\alpha}(\theta^{\prime}-\theta)^2\right) \\
&=\exp \left(\frac{L D^2}{4}\right) Z(\theta)
\end{aligned}
$$
Following the same argument we can show
$$
\tilde{Z}(\theta) \geq \exp \left(\frac{-L D^2}{4}\right) Z(\theta).
$$
In this way, we get bounds between the normalizing constants of original proposal and our approximated proposal.

\subsection{Inequalities of Minimums}
We show $\tilde{Q}\left(\theta^{\prime}, \theta\right) \geq c \cdot Q\left(\theta^{\prime}, \theta\right)$ for $c=\exp \left(\frac{-L D^2}{2}\right)$.
Since both $Q\left(\theta^{\prime}, \theta\right)=\min \{a, b\}$ and $\tilde{Q}\left(\theta^{\prime}, \theta\right)=\min \left\{\tilde{a}, \tilde{b}\right\}$, it is sufficient to show $\tilde{a} \geq c \cdot a$ and $\tilde{b} \geq c \cdot b$ to prove the desired result.
We begin with the $a$ terms
$$
\begin{aligned}
\frac{\tilde{a}}{a} &=\frac{\exp \left(\frac{1}{2}\tilde{\Delta}\left(\theta^{\prime}, \theta\right)-\frac{1}{2\alpha}(\theta^{\prime}-\theta)^2\right)}{\tilde{Z}(\theta)}
\frac{Z(\theta)}{\exp \left(\frac{1}{2}\Delta\left(\theta^{\prime}, \theta\right)-\frac{1}{2\alpha}(\theta^{\prime}-\theta)^2\right)}\\
&=\frac{Z(\theta)}{\tilde{Z}(\theta)} \exp \left(\frac{1}{2}{\tilde{\Delta}\left(\theta^{\prime}, \theta\right)}-\frac{1}{2}\Delta\left(\theta^{\prime}, \theta\right)\right) \\
& \geq \exp \left(\frac{-L D^2}{4}\right) \exp \left(\frac{1}{2}{\tilde{\Delta}\left(\theta^{\prime}, \theta\right)}-\frac{1}{2}\Delta\left(\theta^{\prime}, \theta\right)\right) \\
& \geq \exp \left(\frac{-L D^2}{4}\right) \exp \left(\frac{-L D^2}{4}\right) \\
&=\exp \left(\frac{-L D^2}{2}\right)
\end{aligned}
$$

Now the $b$ terms
$$
\begin{aligned}
\frac{\tilde{b}}{b} &=\frac{\exp\left(\Delta(\theta^{\prime},\theta)+\frac{1}{2}\tilde{\Delta}(\theta,\theta^{\prime})-\frac{1}{2\alpha}(\theta^{\prime}-\theta)^2\right)}{\tilde{Z}(\theta)^{\prime}}
\frac{Z\left(\theta^{\prime}\right)}{\exp \left(\frac{1}{2}\Delta\left(\theta^{\prime}, \theta\right)-\frac{1}{2\alpha}(\theta^{\prime}-\theta)^2\right)}\\
&=\frac{Z\left(\theta^{\prime}\right)}{\tilde{Z}\left(\theta^{\prime}\right)} \frac{\exp \left(\Delta\left(\theta^{\prime}, \theta\right)+\frac{1}{2}\tilde{\Delta}\left(\theta, \theta^{\prime}\right)\right)}{\exp \left(\frac{1}{2}\Delta\left(\theta^{\prime}, \theta\right)\right)} \\
&=\frac{Z\left(x^{\prime}\right)}{\tilde{Z}\left(x^{\prime}\right)} \exp \left(\frac{1}{2}\Delta\left(\theta^{\prime}, \theta\right)+\frac{1}{2}\tilde{\Delta}\left(\theta, \theta^{\prime}\right)\right) \\
& \geq \exp \left(\frac{-L D^2}{4}\right) \exp \left(\frac{1}{2}\Delta\left(\theta^{\prime}, \theta\right)+\frac{1}{2}\tilde{\Delta}\left(\theta, \theta^{\prime}\right)\right)\\
&=\exp \left(\frac{-L D^2}{4}\right) \exp \left(\frac{1}{2}{\tilde{\Delta}\left(\theta, \theta^{\prime}\right)}-\frac{1}{2}\Delta\left(\theta, \theta^{\prime}\right)\right) \\
& \geq \exp \left(\frac{-L D^2}{4}\right) \exp \left(\frac{-L D^2}{4}\right) \\
&=\exp \left(\frac{-L D^2}{2}\right)
\end{aligned}
$$

\subsection{Conclusions}
We have shown that $\tilde{a} \geq \exp \left(\frac{-L D^2}{2}\right) a$ and $\tilde{b} \geq \exp \left(\frac{-L D^2}{2}\right) b$ and therefore it holds that
$$
\tilde{Q}\left(\theta^{\prime}, \theta\right) \geq \exp \left(\frac{-L D^2}{2}\right) Q\left(\theta^{\prime}, \theta\right)
$$
From this, the main result follows directly from Theorem 2 of \citep{zanella2020informed}.
\end{proof}

\section{Experiments on Ising Model}
In the main body of the paper, we have shown that for distributions without
natural differentiable extension, our Newton proposal outperforms popular baselines, including Gibbs sampler and locally-balanced sampler (LB)~\citep{zanella2020informed}, while continuous relaxation-based proposals become valid.
We also apply Newton proposal to discrete distributions with natural differentiable distributions, such as the Ising model, to show the broad applicability of our method. In this case, we compare our Newton proposal with proposals which do not rely on gradients~(Gibbs sampler and LB) and continuous relaxation methods (GWG\citep{grathwohl2021oops} and DLP~\citep{zhang2022langevin}).

\subsection{Experiment Settings}
We have shown in Section 5 that Newton proposal is equivalent to DLP when the discrete distribution has a natural differential extension.
Therefore, if Newton proposal and DLP are applied to sample from the same discrete distribution, they will have the same results.
However, we also demonstrate in Theorem 1 that the smallest eigenvalue of $A$ is related to the asymptotic convergence.
Consider the Ising model whose distribution is
\begin{equation}\label{eq:isingm}
    f(D) = \sum_{u,v\in D}A(u,v)+\sum_{u\in D} b(u),
\end{equation}
where $A$ is a binary adjacency matrix, $a$ is the connectivity strength and $b$ is the bias. 
We can see the diagonal of matrix $A$ in Equation \eqref{eq:isingm} will not affect the distribution in the discrete domain because $u$ and $v$ are different elements in the set $D$.
Meanwhile, DLP is applied to a discrete distribution in a 0-diagonal quadratic form, which can be seen as the multilinear extension of the distribution of Newton proposal when $A$ is 0-diagonal.
We are interested in the performance of Newton proposal and DLP when $\lambda_{min}$ in Newton proposal is larger than that in DLP.

\subsection{Ising Model Sampling Results}
We consider a 4 by 4 lattice Ising model with random variable $\theta \in\{-1,1\}^{d}$, and $d=4 \times 4=16$. 
The distribution is
\[
f(D) = \sum_{u,v\in D}A(u,v)+\sum_{u\in D} b(u)
\]
where $A$ is a binary adjacency matrix, $a=0.1$ is the connectivity strength and $b=0.2$ is the bias. 
\begin{figure}[t]
    \centering
    \subfigure[$\log{\text{RMSE}}$ w.r.t. Iterations]{
        \includegraphics[height=5.2cm]{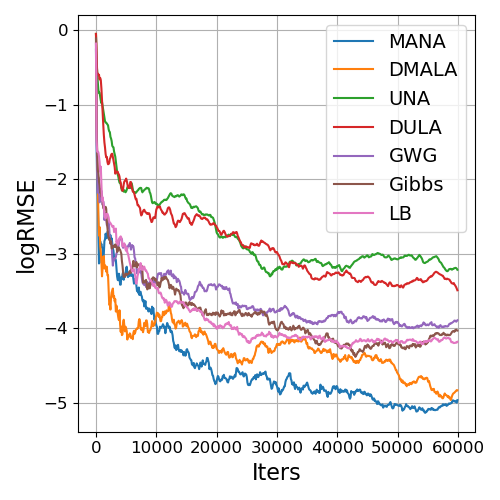}\label{fig:ising1}
    }
    \subfigure[$\log{\text{RMSE}}$ w.r.t. Runtime]{
        \includegraphics[height=5.2cm]{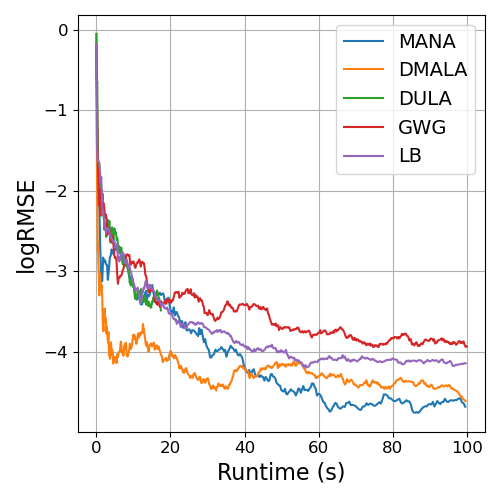}\label{fig:ising2}
    }
    \subfigure[ESS of Proposals]{
        \includegraphics[height=5.2cm]{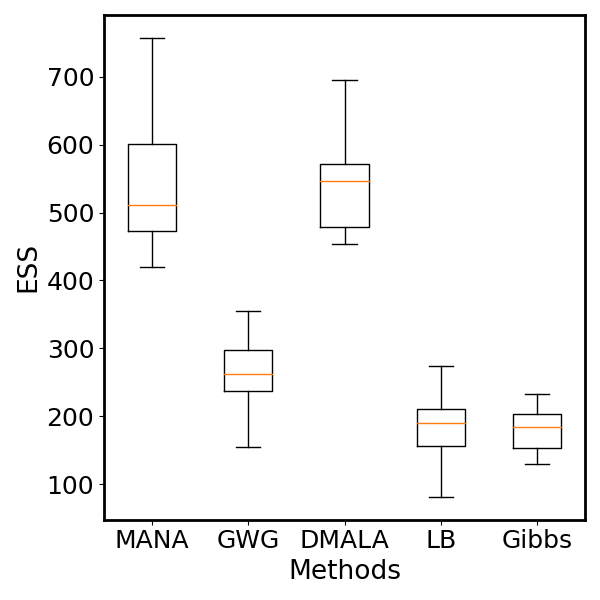}\label{fig:ess}
    }
    \caption{Ising model sampling results. \textbf{(a)} MANA converges faster than the baselines in number of iterations.
    \textbf{(b)} MANA converges faster than the baselines in the same running time.
    \textbf{(c)} MANA and DMALA yield the largest effective sample size (ESS) among all the methods compared.}
    \label{fig:ising}
\end{figure}

We run 60000 iterations with all samplers. To make the comparison of convergence speed fair, we tune the stepsizes so that MANA and DMALA change almost the same number of coordinates in a single update.
The stepsizes of MANA, DMALA, UNA and DULA as 0.5, 0.8, 0.1 and 0.1, respectively.

We first compare the root-mean-square error (RMSE) between the estimated mean and the true mean in Figure \ref{fig:ising}.
MANA is the fastest to converge in terms of both iterations and runtime.
This demonstrates (1) changing many coordinates in one step accelerates the convergence compared to LB and GWG; (2) the finite difference works like the gradient in cases with natural differential extensions to explore the discrete space.
In fact, the Newton proposal is equivalent to DLP when the discrete distribution has a natural differential extension, as shown in Section 5.
That's why DMALA and MANA achieve similar convergence when they change the same number of coordinates in a single update.
MANA and DMALA converge obviously faster than UNA and DULA because the MH correction accelerates the convergence for this task.
In Figure \ref{fig:ess}, we compare the effective sample size (ESS) of different samplers. 
MANA and DMALA significantly outperform other methods, indicating the correlation among its samples is low due to making significant updates in each step.

\section{ROUGE Score in Text Summarization }
We used the official implementation of Rouge score evaluation toolkit~(https://pypi.org/project/rouge-score/).
We follow the same ROUGE score configuration as the settings in \citep{lin2004rouge}: ROUGE version 1.5.5 with options: -a -c 95 -b 665 -m -n 4 -w 1.2.

\section{Dimension-wise Majority Vote in Image Retrieval}
For the image retrieval task, the pseudo code for the dimension-wise majority vote algorithm mentioned in the main body of our paper is as follows:
\begin{algorithm}
\caption{Dimension-wise Majority Vote}
\begin{algorithmic}[0]
\STATE $X\in\{0, 1\}^{K, D}$
\STATE ans $\gets$ zeros([D])
\STATE indices $\gets$ argsort(mean($X$))
\STATE $\triangleright$ select top dimensions under cost constraint $C$
\STATE $\text{ans}[\text{indices}[\colon C]] = 1$
\STATE return ans
\end{algorithmic}
\end{algorithm}

\end{document}